\documentclass[11pt]{article}

\usepackage[preprint]{acl}

\usepackage{times}
\usepackage{latexsym}
\usepackage{amsmath}
\usepackage{amssymb}
\usepackage{booktabs}
\usepackage{multirow}
\usepackage{enumitem}

\usepackage[T1]{fontenc}
\usepackage[utf8]{inputenc}
\usepackage{todonotes}
\usepackage{microtype}
\usepackage{inconsolata}
\usepackage{graphicx}
\PassOptionsToPackage{hyphens}{url}
\usepackage{fvextra}
\usepackage{xcolor}

\fvset{frame=none, numbers=none, fontsize=\scriptsize, baselinestretch=0.92,
       breaklines=true, breakanywhere=true, breakautoindent=false,
       breaksymbolleft=\textcolor{gray}{\tiny$\hookrightarrow$},
       breaksymbolright=\textcolor{gray}{\tiny$\hookrightarrow$}}

\title{The Analyst in the Prompt:\\ Role, Retrieval, and Memory Biases in LLM Financial Analysis}

\author{
Ahmed Asaad\thanks{Correspondence: \texttt{{ahmed.a.abdelazeem@durham.ac.uk}}} \\
Durham University Business School
\And
Amr Mohamed \\
MBZUAI \& Ecole Polytechnique
\AND
Yang Zhang \\
Ecole Polytechnique
\And
Omneya Abdelsalam \\
HBKU \&
Durham University Business School
}

\begin{document}
\maketitle

% Set PDF metadata after \maketitle so it overrides the acl.sty template default.
\hypersetup{
  pdftitle={The Analyst in the Prompt: Role, Retrieval, and Memory Biases in LLM Financial Analysis},
  pdfauthor={Ahmed Asaad, Amr Mohamed, Yang Zhang, Omneya Abdelsalam}
}

\begin{abstract}
Large Language Models (LLMs) increasingly use user context such as memory, profiles, and role prompts to personalize their responses. This personalization can affect evidence-based judgment: the same evidence may lead to different conclusions under different user contexts. Finance provides a high-stakes setting to study this problem because decisions often depend on interpreting long and complex documents. We test this using $3{,}575$ SEC filings across twelve LLMs. We compare persona-conditioned retrieval, neutral retrieval, and memory-framed context to separate the effect of evidence selection from the effect of interpretation. We find that most user-context spillover comes from how models interpret the same evidence under different roles, rather than from retrieving different evidence. We then test two simple mitigation strategies: expressing the same investor mindset as a user profile instead of an assistant role, and separating evidence-based and personalized outputs. Both reduce spillover, but neither removes it completely, and their effectiveness varies substantially across models.

\begin{figure*}[!t]
\centering
\includegraphics[width=\textwidth]{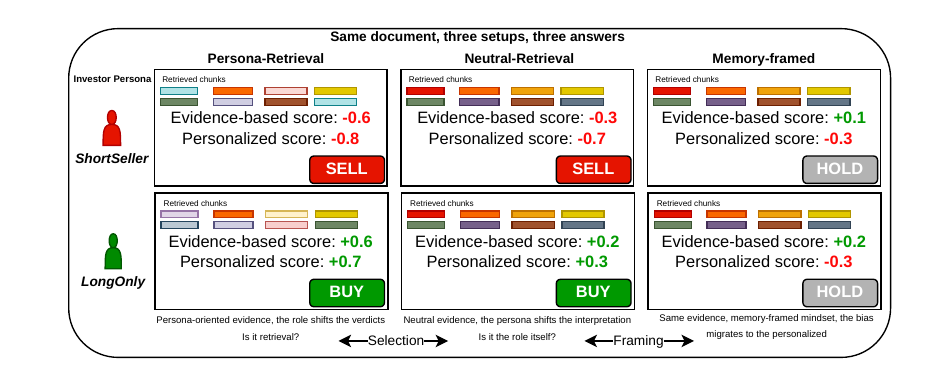}
\caption{Three audit conditions on one Coca-Cola filing.
% [Rebuttal: Reviewer JrEK Comment 3 -- define the two scores in the figure]
The model emits two scores in $[-1,+1]$: the \texttt{EvidenceScore}, a neutral reading of the retrieved evidence that should be invariant to user context (negative${}={}$bearish, positive${}={}$bullish), and the \texttt{PersonalizedScore}, which is allowed to adapt to the user. \textbf{PR}: the persona affects retrieval and interpretation, so personas may read different chunks. \textbf{NR}: personas keep their role prompts but read the same neutral-retrieved chunks, isolating interpretation under fixed evidence. \textbf{MEM (role-to-profile reframing)}: the same chunks are read by a neutral analyst, with the mindset supplied as a user-memory profile rather than as the model's role. Left to right, the figure shows role-shaped verdicts, role-shaped interpretation of identical evidence, and routing of the user signal into the personalized field.}
\label{fig:cocacola}
\end{figure*}
\end{abstract}

% ---------------------------------------------------------------------------
% Main body
% ---------------------------------------------------------------------------
\section{Introduction}
\label{sec:intro}

Large language models (LLMs) are increasingly adapted to users through interaction histories, declared profiles, and role prompts \citep{salemi2024lamp,zhang2024guided,kim2024persona}. Personalization is useful, but it creates a stability problem: desirable adaptation can be hard to distinguish from user-conditioned bias \citep{kantharuban2025stereotype}. The same document, evidence, and factual question can yield systematically distinct \emph{evidence-based} judgments depending on user context. Unlike one-off prompt perturbations, persistent context can affect many future judgments, making this an LLM reliability concern.

This matters for NLP because deployed systems increasingly combine retrieval, long-context reasoning, memory, and user profiles. Standard evaluation asks whether an answer is useful, faithful, or personalized; we ask a sharper invariance question: when a field is defined as a neutral evidence-based judgment, does user context stay out of it? A system can cite the right evidence and still be unstable if the same evidence is interpreted differently across user contexts. We treat finance as a controlled instance of a broader generation problem: context-conditioned bias, where role labels, personality-like traits, user profiles, or stated goals may shift a generated judgment even when the task asks for an invariant, evidence-based answer.

Finance is a natural setting for operationalizing this broader problem in a controlled way. Corporate financial reports are long, information-dense documents used in financial NLP to study risk, sentiment, numerical reasoning, and market-relevant disclosure \citep{kogan2009predicting,loughran2011liability,chen2021finqa}. Investor personas provide concrete, domain-grounded operationalization of role and personality-like context: they encode traits and known investor goals such as risk tolerance, diversification, value or growth exposure, and momentum trading \citep{markowitz1952portfolio,sharpe1964capital,fama1992cross,jegadeesh1993returns}. These objectives may shape a personalized recommendation, but not a neutral reading of the same evidence. As LLM trading systems emerge, such bias becomes more critical.

% \paragraph{Non-technical summary.}
We ask whether a model gives consistent evidence-based judgment when the document is fixed but user context changes. If the neutral judgment moves, the shift could come from retrieving different evidence, interpreting the same evidence differently, or changing how user context is framed. We isolate these channels across $3{,}575$ SEC 10-K/10-Q filings\footnote{SEC Form 10-K is a U.S. public company's annual report; Form 10-Q is its quarterly report. Both contain financial statements, management discussion, and risk disclosures.} and twelve models from five families. The central test is whether personalization changes what the evidence says in its neutral score, rather than how it adapts advice to the user.

Different users want different things, expressed as roles, profiles, traits, preferences, or economic goals. Our target is narrower: a neutral evidence score that should not change because such context is present. We call it user-context spillover when the user signal enters the neutral evidence score. To locate this, we use three matched conditions that vary one part of the pipeline at a time: whether user context affects retrieval, whether it affects interpretation of fixed evidence, and whether the same mindset is framed as the assistant's role or as user memory. This separates evidence selection, interpretation under fixed evidence, and residual spillover into the neutral evidence-based score. The design treats personalization as appropriate only when routed to a separate user-adapted output, rather than allowing it to shift the evidence-only judgment; this distinction provides a test of user-context invariance in long-document analysis.

This paper makes the following contributions:
\textbf{(1)} a three-condition audit design for measuring where role-, profile-,
trait-, or goal-conditioned context enters an LLM generation pipeline in a
long-document setting;
\textbf{(2)} a cross-model audit showing that the main failure is
\emph{interpretive drift}: models judge the same filing differently under
different roles;
\textbf{(3)} evidence that prompt placement matters: the same mindset causes
less spillover when passed as user context than when assigned as the assistant's
role; and
\textbf{(4)} an analysis showing that the effects are systematic: they flip recommendations, track sector-level language, distort risk signals, and vary across document populations.
\section{Related Work}
\label{sec:related}

\paragraph{Financial NLP and finance-LLM evaluation.}
Financial NLP has long shown that language, disclosure structure, and economic context matter for financial text. Earlier work links financial language to sentiment, risk, market outcomes, and firm fundamentals \citep{tetlock2007giving,loughran2011liability,kogan2009predicting,malo2014financialphrasebank}. Recent finance-oriented models and benchmarks evaluate sentiment, numerical reasoning, forecasting, knowledge, and decision support \citep{araci2019finbert,chen2021finqa,chen2022convfinqa,wu2023bloomberggpt,yang2023fingpt,xie2023pixiu,matlin2025flame,klimaszewski2025avenibench,shah2025financialknowledge}. This work measures task performance. We ask whether evidence-based judgments remain stable when the filing is fixed but user context changes.
\paragraph{Prompt sensitivity, personas, and persistent context.}
LLM outputs can shift under prompt changes, including formatting, wording, adversarial variants, and prompt-sensitivity benchmarks \citep{sclar2024promptformatting,zhuo2024prosa,zhu2023promptrobust,razavi2025promptsensitivity}. Persona and role prompts are a relevant source of variation: identities and system prompt position can alter model responses \citep{kim2024persona,lutz2025persona,neumann2025position}. We build on this, but move from observing sensitivity to decomposing where it enters: evidence retrieval, interpretation of fixed evidence, or framing of persistent user context.

\paragraph{Bias and financial decision aids.}
Work on LLM bias shows that political and entity-level priors can surface even when irrelevant to the task. Finance-specific studies document scenario priors, role anchoring, look-ahead leakage, foreign bias, and entity-level memorization in financial LLMs \citep{zhou2025rationalinvestors,lee2025yourai,liu2026scenario,biancotti2025chatbankmanfried,cao2025foreignbias,kong2026biasfinance}. LLMs are proposed as assistants for investment analysis, trading, credit assessment, and multi-agent financial workflows \citep{xiao2025tradingagents,drinkall2025creditratings}. If role prompts shift neutral conclusions independently of evidence, they introduce an unmeasured source of model risk.

% \paragraph{External validation of prompt-induced judgments.}
% Language-derived signals are known to relate to market outcomes such as returns and volatility \citep{tetlock2007giving,loughran2011liability,kogan2009predicting,lopezlira2023chatgpt,kim2024financialstatement}. We use this literature narrowly: after estimating prompt-induced shifts under fixed filings, we test whether the resulting scores relate to future post-filing returns (Appendix~\ref{app:extra}).
\section{Methodology}
\label{sec:design}
We design the study to separate three potential sources of user-context effects in LLM financial analysis: evidence selection, interpretation of fixed evidence, and the framing of user preferences. We first define the filing-analysis task and two-score output, then introduce the three experimental conditions used to isolate these channels. We next describe the retrieval pipeline and SEC filing sample, followed by the filing-fixed-effects regressions used to estimate persona and memory-profile effects. Finally, we define the headline metrics and decomposition used to quantify retention, framing shrinkage, and residual spillover across conditions.

\subsection{Task and two-score output}

Given an SEC 10-K or 10-Q filing reduced to MD\&A, Risk Factors, and Quantitative and Qualitative Disclosures\footnote{These correspond to the main 10-K/10-Q textual disclosure sections commonly studied in prior work on filing narratives, including \citet{cohen2020lazyprices}.}, the model returns a fixed JSON schema with two scores in $[-1,+1]$: an \texttt{EvidenceScore}, intended to reflect a neutral reading of the retrieved chunks, and a \texttt{PersonalizedScore}, allowed to reflect the prompted role or user-memory profile. The schema also includes rationales and auxiliary fields, but is identical in all conditions; only the prompt context changes. We call role- or user-context movement in \texttt{EvidenceScore} \emph{user-context spillover}.
% [Rebuttal: Reviewer JrEK W4 -- bias is movement vs a reference, not deviation from ground truth]
Throughout, the neutral condition is an \emph{invariance reference}, not an externally validated ground-truth judgment: spillover measures how far the evidence-based score \emph{moves} when user context changes, not that the moved score is wrong. Because every metric is a within-filing difference across conditions, any constant error in the neutral reading cancels. Movement in \texttt{PersonalizedScore} is expected, because that field is explicitly allowed to adapt to the user.
% [Rebuttal: Reviewer JrEK Comment 4 -- terminology consistency]
We use the following terms consistently: a \emph{persona} is an investor mindset; an \emph{assistant role} places that mindset in the system prompt (``you are a short seller''); a \emph{user profile} / \emph{user-memory profile} places the same mindset as third-person user context (``the user is a downside-focused investor''); and \emph{user-context spillover} is the resulting movement of the neutral \texttt{EvidenceScore}.

\subsection{Three conditions, one lever per arm}

We use nine investor personas plus a neutral baseline, spanning common viewpoints such
as short selling, long-only investing, ESG, momentum, credit, value, retail, and
sell-side analysis. The memory setup uses five matched user-memory profiles for the
strongest-signal roles; full blurbs appear in Appendix~\ref{app:prompts}. The headline decomposition in Figure~\ref{fig:decomp_per_model} and
Table~\ref{tab:crossmodel} averages over the four strongest-signal matched pairs:
\emph{ShortSeller--DownsideProtection}, \emph{RiskManager--RiskAverse},
\emph{LongOnly--LongOnlyGrowth}, and \emph{ESG--ESGFocused}. The fifth pair,
\emph{Momentum--MomentumTrader}, has a near-zero base effect and appears per model
in the cross-model panel, Figure~\ref{fig:crossmodel}, but is excluded from the headline metrics.

\noindent $\bullet$ \textbf{Persona Retrieval (PR).} The persona conditions both the retrieval query and the model prompt, so personas may retrieve various chunks.

\noindent $\bullet$ \textbf{Neutral Retrieval (NR).} Retrieval uses a neutral query once per filing, and the resulting chunk set is reused verbatim across personas; any surviving persona effect is therefore generation-stage.

\noindent $\bullet$ \textbf{Memory with Neutral Retrieval (MEM).} Retrieval remains neutral and pinned: every MEM call on a filing reads the same top-$k$ chunks as the NR calls. The role block is reset to the neutral analyst, and a new user context line carries the matched memory profile in the third person. For example, \texttt{RiskManager} is replaced by a \texttt{RiskAverse} profile: the assistant no longer acts as a risk manager but serves a user prioritizing capital preservation and low drawdown tolerance, simulating prior user interactions stored as persistent memory.
\subsection{Retrieval and sample}

Each filing is chunked into approximately $400$-token segments with $80$-token overlap, embedded once with OpenAI \texttt{text-embedding-3-large} (dim $3072$), and indexed per filing. Retrieval returns the top-$k=8$ chunks.\footnote{We treat $k$ as a fixed evidence budget rather than a tuned performance parameter; $8$ is close to the $k=5$--$10$ range used in early retrieval-augmented generation experiments \citep{lewis2020rag}.} We use a stratified sample of $3{,}575$ SEC filings from firms in the S\&P 500\footnote{The S\&P 500 is a large-cap U.S. equity index commonly used as a benchmark for major publicly traded firms. The universe is firms that were index constituents at some point during the sample period, so the number of distinct firms exceeds $500$ because of index turnover.}, maintaining the index's GICS\footnote{GICS, the Global Industry Classification Standard, groups firms into sectors such as Information Technology, Health Care, Financials, and Energy.} sector distribution across filing years $2013$--$2024$, with at most six filings per firm to keep the cross-section firm-diversified. All calls use deterministic decoding ($T=0$, top-$p=1$, max-tokens $1024$). Retrieval construction and byte-identical retrieval checks are detailed in Appendix~\ref{app:reproducibility}, while exact retrieval queries are reproduced in Appendix~\ref{app:retr_query}.

\subsection{Experiments}
\label{sec:regressions}

For each model, we fit six filing-fixed-effects regressions: the three conditions (PR, NR, MEM) crossed with the two score fields, \texttt{EvidenceScore} and \texttt{PersonalizedScore}. Each regression has the form
\begin{equation}
s_{i,f} = \alpha_i + \boldsymbol\beta^{\top} \mathbf{z}_f + \varepsilon_{i,f},
\label{eq:fereg}
\end{equation}
where $i$ indexes the filing, $f$ indexes the frame (persona in PR/NR and profile in MEM), $\alpha_i$ is the filing fixed effect, and $\mathbf{z}_f$ is a one-hot vector for non-neutral frames. The neutral frame is the reference, so each coefficient is the marginal effect of one persona or profile on one score field, holding the filing fixed. Standard errors are clustered at the filing level, with $95\%$ confidence intervals computed using a bootstrap ($B = 1{,}000$ replications); further regression details appear in Appendix~\ref{app:regression}.

\subsection{Metrics and decomposition}
\label{sec:metrics}

The headline metrics use four coefficients from the filing-fixed-effect regressions of \S\ref{sec:regressions}. Let $\hat\beta_p^{\mathrm{PR}}$ and $\hat\beta_p^{\mathrm{NR}}$ denote the effect of persona $p$ on the \emph{evidence-based} score under PR and NR. Let $\hat\gamma_m^{\mathrm{ev}}$ denote the effect of profile $m$ on the evidence-based score under MEM, and $\hat\delta_m^{\mathrm{pers}}$ its effect on the \emph{personalized} score. We use absolute values since personas may move scores in opposite directions; the metrics measure the magnitude of spillover, not its sign.
\begin{align}
\mathrm{Bias}_p
&= \left| \hat\beta_p^{\,\mathrm{PR}} \right| ,
\label{eq:bias}
\end{align}
\begin{align}
\mathrm{Retain}_p
&=
\frac{\left| \hat\beta_p^{\,\mathrm{NR}} \right|}
     {\left| \hat\beta_p^{\,\mathrm{PR}} \right|},
\label{eq:retain}
\end{align}
\begin{align}
\mathrm{Shrink}(p,m)
&=
1 - \frac{\left| \hat\gamma_m^{\,\mathrm{ev}} \right|}
         {\left| \hat\beta_p^{\,\mathrm{NR}} \right|},
\label{eq:shrink}
\end{align}
\begin{align}
L_m
&=
\frac{\left| \hat\gamma_m^{\,\mathrm{ev}} \right|}
     {\left| \hat\gamma_m^{\,\mathrm{ev}} \right|
      + \left| \hat\delta_m^{\,\mathrm{pers}} \right|}.
\label{eq:leak}
\end{align}
\noindent
\textbf{Bias} is the persona effect under persona-based retrieval. \textbf{Retention} is the share of that effect that survives byte-identical retrieval, isolating interpretation of the same evidence. \textbf{Framing shrinkage} is the share of the NR effect removed when the same mindset is reframed from assistant role to user memory profile; it is defined only for matched persona--profile pairs. \textbf{Leakage} is the share of the MEM user signal that remains in the evidence-based score rather than being routed to the personalized score. Thus, $L_m=0$ means perfect routing, while $L_m=0.5$ means no separation: the signal is split evenly across both fields.

For each matched persona--profile pair $(p,m)$, we decompose the PR effect into three channels:
{\small
\begin{align}
|\hat\beta_p^{\mathrm{PR}}|
=\, & \underbrace{|\hat\beta_p^{\mathrm{PR}}| - |\hat\beta_p^{\mathrm{NR}}|}_{\text{evidence selection}} \nonumber \\
+ & \underbrace{|\hat\beta_p^{\mathrm{NR}}| - |\hat\gamma_m^{\mathrm{ev}}|}_{\text{context framing}}
+ \underbrace{|\hat\gamma_m^{\mathrm{ev}}|}_{\text{residual spillover}}.
\label{eq:decomp}
\end{align}
}

\noindent
Equation~\ref{eq:decomp} splits the original PR effect into the part removed by fixing retrieval, the part removed by changing the framing of the same mindset, and the spillover remaining in the evidence-based score under MEM. When a step increases rather than decreases a cell's magnitude, we set that step's reduction to zero before averaging. Because this clamping only removes negative reductions, the channel components stacked in Figure~\ref{fig:decomp_per_model} can sum to slightly more than the mean $|\hat\beta_p^{\mathrm{PR}}|$ in Table~\ref{tab:crossmodel} when fixing retrieval increases a coefficient; the equality in Eq.~\ref{eq:decomp} holds for the signed terms before clamping. The personalized score is not part of this decomposition; it is used only in $L_m$ to measure whether the user signal is routed to the personalized output or leaks into the evidence-based score.

\section{Results}
\label{sec:results}

\begin{figure}[t]
\centering
\includegraphics[width=\columnwidth]{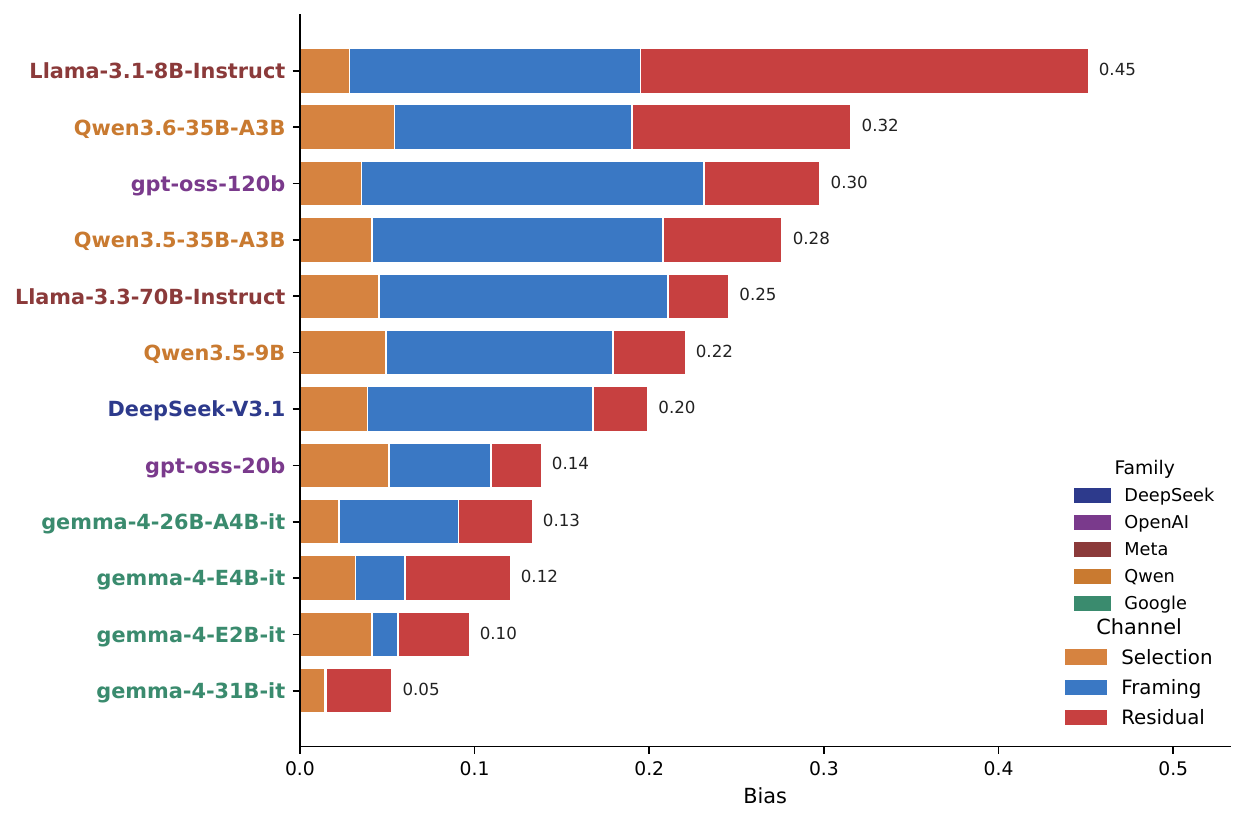}
\caption{Per-model decomposition of role-conditioned bias in the evidence-based score, averaged across the four matched pairs. Each bar stacks the three clamped channel components: evidence selection (orange), prompt framing (blue), and residual spillover (red) that remains after both interventions. The stacked height approximates the mean $|\hat\beta_p^{\mathrm{PR}}|$ in Table~\ref{tab:crossmodel} and can slightly exceed it where fixing retrieval raises a coefficient (clamping; \S\ref{sec:metrics}).}
\label{fig:decomp_per_model}
\end{figure}

\begin{figure*}[t]
\centering
\includegraphics[width=\textwidth]{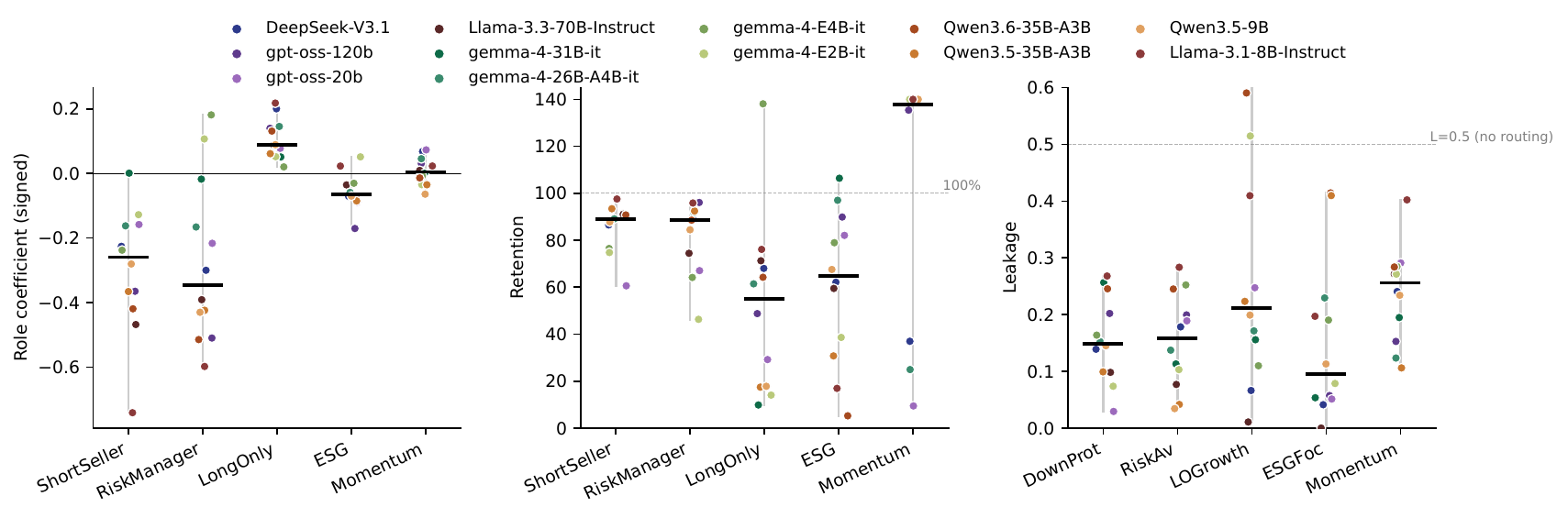}
\caption{Cross-model results across twelve models and $3{,}575$ filings. Each dot is one model--persona/profile estimate; gray bars show the cross-model range and black bars the panel median. \textbf{Left}: the signed per-persona role coefficient $\hat\beta_p^{\mathrm{PR}}$ on the evidence-based score under PR (negative = bearish, positive = bullish); its magnitude is the Bias metric of Eq.~\ref{eq:bias}. \textbf{Middle}: retained bias when all personas read the same evidence; values near $100\%$ indicate interpretation rather than retrieval, and values above $100\%$ mean the neutral-retrieval coefficient exceeds the persona-retrieval one (typically for weak base effects). \textbf{Right}: MEM leakage, measuring user signal left in the evidence-based score rather than routed to the personalized score; $L=0$ is perfect routing and $L=0.5$ is no routing.}
\label{fig:crossmodel}
\end{figure*}

We report results on the \emph{evidence-based} score, the field the model was instructed to keep invariant to role or user context. The personalized-score coefficients enter only through the leakage ratio $L_m$ (\S\ref{sec:personalisation_routing}). Figure~\ref{fig:decomp_per_model} summarizes the three-channel decomposition by model. Figure~\ref{fig:crossmodel} shows the cross-model spread on the headline metrics, and Table~\ref{tab:crossmodel} reports the corresponding per-model averages.

\subsection{Most bias survives identical retrieval}
\label{sec:evidence_selection}

We first test whether role-conditioned bias is primarily an evidence-selection effect. In PR, the role conditions both retrieval and generation, so personas may read different chunks from the same filing. In NR, retrieval is neutral and pinned: every persona receives the same byte-identical chunk set for a filing. If retrieval were the main driver, the NR effect should fall sharply toward zero. Instead, it largely persists. Across the four matched pairs, the panel retains $69\%$ of the PR effect on average and $68\%$ at the median, with model-level retention ranging from $44$--$89\%$ (Table~\ref{tab:crossmodel}). The strongest direction, \texttt{ShortSeller}, is even more persistent: eight of twelve models retain at least $74\%$ of the PR effect, and the panel median is $89.1\%$. Thus, under this retrieval configuration, role-conditioned movement in the evidence-based score is not primarily explained by retrieval drift. The dominant source is interpretive drift: models assign different evidence-based scores to the same filing chunks once a role is in scope. This pattern is consistent across model families and parameter scales. The magnitude of bias varies substantially, but does not decrease monotonically with scale: larger models are not uniformly more invariant. The more stable regularity is the channel structure. Evidence selection is usually a minority component, while prompt framing and residual spillover account for most of the total effect (Figure~\ref{fig:decomp_per_model}).

\begin{table}[t]
\centering
\small
\setlength{\tabcolsep}{3pt}
\renewcommand{\arraystretch}{0.95}
\begin{tabular}{lrrr}
\toprule
Model & $\overline{|\hat\beta_p^{\text{PR}}|}$ & $\mathrm{Retain}_p$ & $\overline{L_m}$ \\\midrule
DeepSeek-V3.1       & $0.20$ & $76\%$ & $0.11$ \\
gpt-oss-120b        & $0.30$ & $82\%$ & $0.34$ \\
Llama-3.3-70B-Instruct       & $0.25$ & $74\%$ & $0.05$ \\
Qwen3.6-35B-A3B & $0.29$ & $62\%$ & $0.37$ \\
Qwen3.5-35B-A3B & $0.23$ & $59\%$ & $0.19$ \\
gemma-4-31B-it         & $0.02$ & $58\%$ & $0.15$ \\
gemma-4-26B-A4B-it     & $0.13$ & $85\%$ & $0.17$ \\
gpt-oss-20b         & $0.14$ & $60\%$ & $0.13$ \\
Qwen3.5-9B          & $0.22$ & $64\%$ & $0.12$ \\
Llama-3.1-8B-Instruct        & $0.40$ & $72\%$ & $0.29$ \\
gemma-4-E4B-it     & $0.12$ & $89\%$ & $0.18$ \\
gemma-4-E2B-it      & $0.08$ & $44\%$ & $0.19$ \\
\midrule
\textbf{Panel mean}    & $\mathbf{0.20}$ & $\mathbf{69\%}$ & $\mathbf{0.19}$ \\
\textbf{Panel median}  & $\mathbf{0.21}$ & $\mathbf{68\%}$ & $\mathbf{0.18}$ \\
\textbf{Range}         & $0.02$--$0.40$  & $44$--$89\%$    & $0.05$--$0.37$ \\
\bottomrule
\end{tabular}
\caption{Per-model and panel-aggregate summary, averaged over the four headline (persona, profile) pairs. $\overline{|\hat\beta_p^{\text{PR}}|}$ is on the \textbf{evidence-based} score; \emph{retain} is the ratio of the NR and PR coefficients on the evidence-based score; $\overline{L_m}$ combines the MEM evidence-based and MEM personalized coefficients.}
\label{tab:crossmodel}
\end{table}

\subsection{Prompt framing is the strongest lever}
\label{sec:context_framing}

We next compare NR with MEM. Both conditions hold retrieval fixed, but differ in where the same investor mindset appears. In NR, the mindset is assigned as the assistant's role; in MEM, the assistant is neutral and the mindset is passed as a user memory profile.

This reframing reduces the evidence-side coefficient in the large majority of cells (Appendix~\ref{app:full_bias} catalogues the few weak-signal exceptions). For the strongest matched pair, \texttt{ShortSeller}/\texttt{DownsideProtection}, NR-to-MEM shrinkage ranges from $51\%$ to $87\%$, with a panel median of $73\%$. Across the audit, the same preference signal has a stronger effect when it defines who the assistant is than when it describes the user being served. Prompt placement is not cosmetic; it is a major predictor of whether user context enters the neutral field.

\subsection{Dual-output routing helps}
\label{sec:personalisation_routing}

Under MEM, the assistant is neutral and the user profile is present. The leakage ratio $L_m$ measures how much user signal remains in the evidence-based score rather than being routed to the personalized score: $L_m=0$ is perfect routing, while $L_m=0.5$ means no separation. All twelve models have mean leakage below $0.5$, so the dual-output schema reduces leakage throughout, but never removes it: it is a risk-reduction technique, not a guarantee.
% [Rebuttal: Reviewer pZ9F W2 -- foreground model-dependence; name the leakiest models]
Effectiveness varies widely, with mean leakage ranging from $0.05$ to $0.37$, and this variation does not track model size or raw spillover magnitude. The leakiest routers are \texttt{Qwen3.6-35B-A3B} ($0.37$), \texttt{gpt-oss-120b} ($0.34$), and \texttt{Llama-3.1-8B-Instruct} ($0.29$): some role-sensitive models route cleanly, while others---including a $120$B model---leave substantial residual signal in the evidence-based field. Output separation is useful, but routing is a model-specific property that must be checked per model rather than assumed from scale.

\subsection{Rationales show the same channels}
\label{sec:rationale_shifts}

The rationale text mirrors the score-level decomposition. Using DeepSeek-V3.1 as a case study, Table~\ref{tab:rationale_shifts} reports two examples in which the \texttt{ShortSeller} evidence-based score changes by $0.50$ between PR and NR. Under PR, the model emphasizes downside-oriented risks. Under NR, the same persona reads byte-identical evidence but gives more weight to growth and operating performance. Under MEM, the evidence-based rationale moves closer to the neutral reading, while the downside-oriented interpretation is routed to the personalized rationale.

\begin{table*}[t]
\centering
\footnotesize
\setlength{\tabcolsep}{4pt}
\renewcommand{\arraystretch}{1.15}
\begin{tabular}{p{0.06\textwidth} p{0.24\textwidth} p{0.24\textwidth} p{0.40\textwidth}}
\toprule
& \textbf{PR / ShortSeller}  & \textbf{NR / ShortSeller}  & \textbf{MEM / DownsideProtection} \\
\midrule
\textbf{AMD} \newline 10-Q Q1 2021 &
\textit{ev$=-0.30$, \textbf{SELL}.} ``Gray market sales distorting demand and margins, potential goodwill impairment from the Xilinx merger, and restrictive debt covenants. These factors suggest downside risk\ldots'' &
\textit{ev$=+0.20$, \textbf{SELL}.} ``Revenue surged $93\%$ year-over-year to \$$3.4$B, driven by strong growth in both segments. However, the filing contains extensive forward-looking statements\ldots'' &
\textit{ev$=+0.30$, pers$=-0.20$, \textbf{HOLD}.} ev: ``Revenue surged $93\%$ year-over-year to \$$3.4$B\ldots adequate liquidity\ldots'' \newline pers: ``\textbf{The user's tail-risk focus} weighs the potential adverse outcomes from tax audits and acquisition integration risks, which could offset the strong revenue growth.'' \\
\midrule
\textbf{SNPS} \newline 10-Q Q1 2024 &
\textit{ev$=-0.30$, SELL.} ``Significant macroeconomic uncertainties, including inflationary pressures, interest rates, and geopolitical tensions, which could negatively impact the semiconductor industry\ldots'' &
\textit{ev$=+0.20$, \textbf{SELL}.} ``Revenue grew $21\%$ to \$$1.6$B, with operating income up $41\%$, but the quarter included an extra week contributing \$$70.5$M in revenue\ldots pending Ansys acquisition\ldots'' &
\textit{ev$=+0.20$, pers$=-0.30$, \textbf{HOLD}.} ev: ``Revenue grew $21\%$, operating income $+41\%$\ldots forward-looking risks including macro uncertainty.'' \newline pers: ``Given the user's tail-risk focus, the extra week's boost to revenue and the numerous risk factors\ldots \textbf{warrant a hold rather than a buy or sell.}'' \\
\bottomrule
\end{tabular}
\caption{Verbatim rationales for two filings (DeepSeek-V3.1, $\Delta\texttt{EvidenceScore}=0.50$ between PR and NR; \textit{ev}=\texttt{EvidenceScore}, \textit{pers}=\texttt{PersonalizedScore}). Under PR, the model emphasizes downside risks; under NR, the same persona reads fixed evidence but emphasizes growth and operating performance. Under MEM, the evidence-based rationale becomes more neutral, while the personalized rationale carries the user's tail-risk preference.}
\label{tab:rationale_shifts}
\end{table*}

We quantify this convergence by embedding every evidence-based rationale from the same model and comparing each non-neutral frame with the neutral-baseline rationale for the same filing. Higher cosine similarity means closer alignment with the neutral analyst's explanation. Median similarity rises from PR ($0.721$) to NR ($0.805$) to MEM ($0.849$): fixing retrieval makes rationales more similar by forcing all personas to discuss the same evidence, and neutralizing the assistant role increases similarity further. Thus, the channel ordering appears in both scores and explanations: user context changes not only the numerical judgment, but also which evidence the model emphasizes.

\begin{figure}[t]
\centering
\includegraphics[width=\columnwidth]{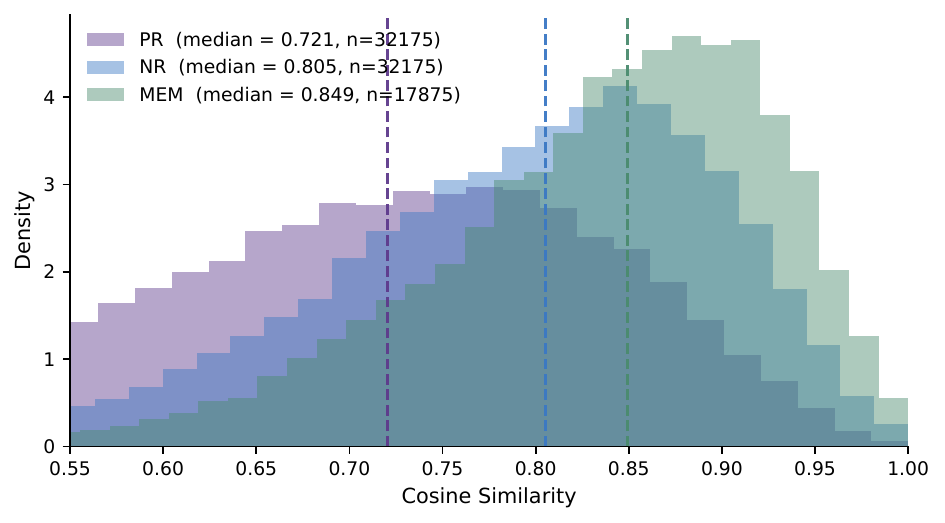}
\caption{Rationale convergence across PR, NR, and MEM: cosine similarity between each non-neutral evidence rationale and the neutral baseline; higher is closer.}
\label{fig:rationale_similarity}
\end{figure}

\begin{figure*}[!t]
\centering
\includegraphics[width=\textwidth]{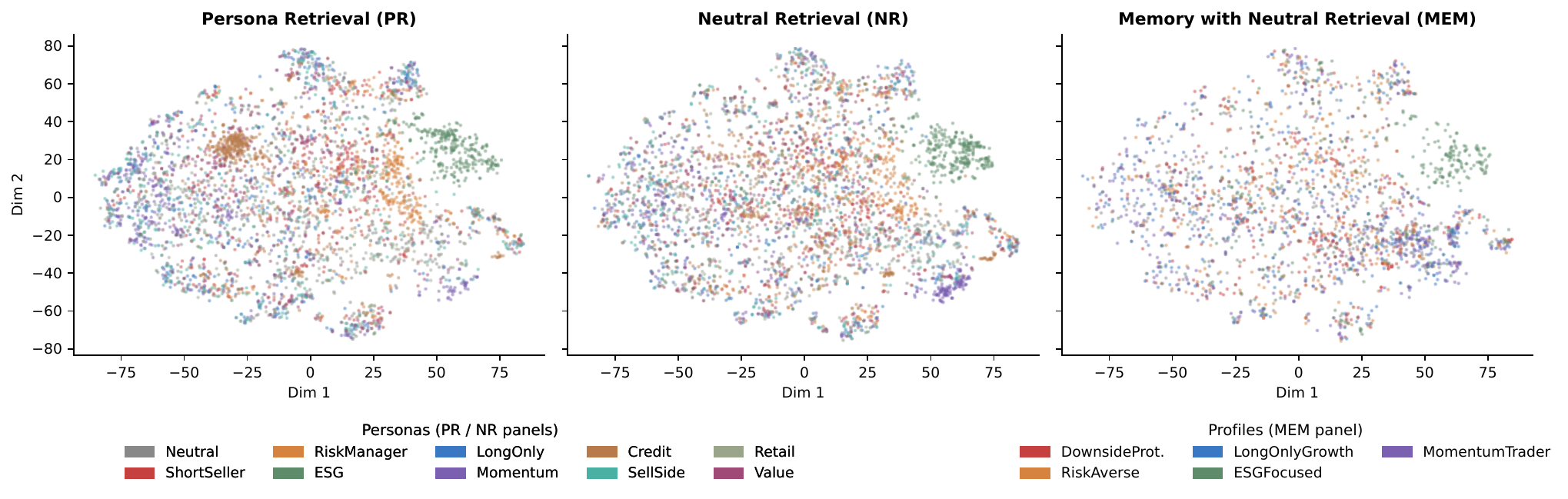}
\caption{Two-dimensional projection of evidence-based rationale embeddings. \textbf{PR}: persona-colored points separate, indicating different rationales under persona-conditioned retrieval and interpretation. \textbf{NR}: points overlap more under byte-identical evidence, although persona-tinted regions remain. \textbf{MEM}: user-profile colors collapse into a near-neutral cloud, showing that rationales converge when the assistant role is neutral.}
\label{fig:rationale_proj}
\end{figure*}

Figure~\ref{fig:rationale_proj} provides a visual check of this result. Each point is one evidence-based rationale, embedded with \texttt{text-embedding-3-large} and projected to two dimensions using PCA-$50$ followed by $t$-SNE on a stratified sample of $\sim 10{,}000$ rationales. The three panels share the same projection, so locations are comparable across conditions. The mean pairwise distance between per-frame centroids decreases from PR to NR to MEM, matching both the cosine-similarity ordering and the score-level decomposition. This structure is not specific to the projection settings: the PR $>$ NR $>$ MEM centroid-spread pattern holds across reasonable $t$-SNE perplexity and PCA-rank choices.

\subsection{Residual leakage varies across document populations}
\label{sec:mcap_bias}

Finally, we test whether residual spillover differs across document populations. We use market capitalization not as a market outcome, but as a proxy for two NLP-relevant properties: disclosure regularity and likely entity familiarity. Splitting filings into market-capitalization buckets reveals a clear gradient. Residual leakage in the evidence-based score is largest on smaller companies with residuals $2$--$4\times$ greater than large-cap residuals across the four headline matched pairs. This suggests that user-context spillover is amplified on less standardized or less familiar documents, where the model may rely more heavily on prompt-conditioned interpretive frames. The result is therefore relevant beyond finance: invariance failures may be uneven across document populations, especially for long-tail entities and less templated texts.

% \subsection{2-D projection of rationale embeddings}
% \label{app:rationale_proj}

\subsection{Do model scores retain signal?}
We ask whether the evidence-based scores contain any forward-looking signal. This analysis is not the main target of the paper: our audit concerns invariance under user context, not trading performance. Still, if the scores were arbitrary, downstream shifts in their level would be less consequential.
For each model--persona pair, we rank filings by the evidence-based score, form quintiles, and compute the cumulative abnormal return of the highest-scored quintile minus the lowest-scored quintile over the year. Returns are measured relative to the S\&P 500. Table~\ref{tab:quintile_by_persona} reports the result for three representative models and the mean across all twelve models. Eight of nine personas show clearly positive top-minus-bottom quintile spreads (overall average $+2.96\%$); only \texttt{RiskManager} is negative for the three representative models, and roughly flat ($+0.08\%$) in the 12-model panel mean.
The relevant conclusion is not that the model scores define a deployable investment strategy. Rather, the within-persona ranking retains signal while persona conditioning shifts score levels and recommendations. This makes the invariance failure practically meaningful: user context can change the apparent evidence-based judgment even when the score field still carries information about future outcomes.

\begin{table*}[h]
\centering
\small
\setlength{\tabcolsep}{4pt}
\begin{tabular*}{\textwidth}{@{\extracolsep{\fill}} lrrrr @{}}
\toprule
Persona & DeepSeek-V3.1 & gpt-oss-120b & Llama-3.3-70B-Instruct & Panel mean (12 models) \\
\midrule
\texttt{Credit}       & $+5.30\%$ & $+3.63\%$ & $+3.07\%$ & $+3.29\%$ \\
\texttt{Value}        & $+4.75\%$ & $+5.14\%$ & $+5.94\%$ & $+5.63\%$ \\
\texttt{SellSide}     & $+4.52\%$ & $+5.22\%$ & $+1.71\%$ & $+3.30\%$ \\
\texttt{LongOnly}     & $+3.68\%$ & $+4.95\%$ & $+3.53\%$ & $+3.84\%$ \\
\texttt{Momentum}     & $+3.60\%$ & $+0.51\%$ & $+2.07\%$ & $+1.70\%$ \\
\texttt{Retail}       & $+3.51\%$ & $+2.83\%$ & $+1.14\%$ & $+2.65\%$ \\
\texttt{ESG}          & $+3.10\%$ & $+3.96\%$ & $+0.72\%$ & $+3.44\%$ \\
\texttt{ShortSeller}  & $+1.04\%$ & $+3.12\%$ & $+0.68\%$ & $+2.72\%$ \\
\texttt{RiskManager}  & $-1.40\%$ & $-0.87\%$ & $-0.45\%$ & $+0.08\%$ \\
\midrule
\textbf{Mean across personas} & $\mathbf{+3.12\%}$ & $\mathbf{+3.17\%}$ & $\mathbf{+2.05\%}$ & $\mathbf{+2.96\%}$ \\
\bottomrule
\end{tabular*}
\caption{Downstream signal in evidence-score rankings, by persona and model. For each model--persona pair, $n \approx 3{,}393$ firm filings (those of the $3{,}575$ with at least $252$ trading days of post-filing prices) are sorted by evidence-based score and split into quintiles. The table reports one-year abnormal return of firms in the highest-scored quintile minus those in the lowest-scored quintile, relative to the S\&P 500. Positive values indicate that firms with higher-scored filings tend to outperform lower-scored firms.}
\label{tab:quintile_by_persona}
\end{table*}
\section{Discussion}
\label{sec:discussion}

\paragraph{Roles shape interpretation, not just evidence selection.}
The main result is an invariance failure in long-document generation: a field explicitly defined as a neutral evidence-based judgment changes when user context changes. Pinning retrieval removes only a minority of the role effect, so the problem is not primarily that different personas retrieve different evidence. Rather, once a role is in scope, models interpret the same evidence differently. The largest reduction we observe comes from changing how the same investor mindset is represented: as the assistant's role, it behaves like an instruction for reading the filing; as user context served by a neutral assistant, its effect on the evidence-side coefficient shrinks substantially. This extends work on prompt sensitivity and persona prompting by showing not only that prompts change outputs, but that role framing can shape interpretation even after retrieval is held fixed \citep{sclar2024promptformatting,zhuo2024prosa,kim2024persona,lutz2025persona}. Investor personas are not arbitrary characters. They correspond to familiar objectives from portfolio choice and asset-pricing theory: risk tolerance, diversification, value/growth exposure, and momentum \citep{markowitz1952portfolio,sharpe1964capital,fama1992cross,jegadeesh1993returns}. Such objectives can legitimately affect a personalized recommendation. The failure we study is narrower: these objectives should not change a score explicitly defined as a neutral reading of the same filing.

\paragraph{Personalization is safer as user context than as assistant identity.}
The role--profile comparison suggests that where a preference enters the prompt is not merely a formatting choice. When the same underlying investor objective defines the assistant's identity, it acts as a lens through which evidence is interpreted; when it instead describes the user served by a neutral assistant, substantially less of that preference enters the evidence-based judgment. Our ablations reinforce this interpretation: the reduction persists even without the dual-output schema, indicating that role-to-profile reframing has an effect distinct from output separation. This suggests a broader design principle for personalized LLM systems: user preferences should be represented as information to condition personalized advice, rather than as instructions that redefine the reasoning identity of the model. Personalization need not be removed; it can instead be routed to outputs where adaptation is intended while preserving greater invariance in outputs meant to reflect the underlying evidence.

\paragraph{Invariance is not implied by scale or task performance.}
The three-channel pattern holds across model families and parameter scales, but the magnitudes do not improve monotonically with size. Larger models are not uniformly safer, and MoE models are not uniformly more invariant. This is consistent with recent finance-LLM evaluations showing that model capability, domain performance, and bias do not necessarily move together \citep{wu2023bloomberggpt,yang2023fingpt,xie2023pixiu,matlin2025flame}. User-context invariance is therefore a model behavior that must be evaluated directly, not inferred from model scale or benchmark performance.

\paragraph{Implications for personalized LLM evaluation.}
Within-persona rankings retain signal: for eight of nine personas, filings in the highest evidence-score quintile outperform those in the lowest quintile over the following trading year after market adjustment. This connects to work showing that language-derived signals relate to market outcomes \citep{tetlock2007giving,loughran2011liability,kogan2009predicting,lopezlira2023chatgpt,kim2024financialstatement}. The key distinction is between \emph{level} and \emph{ranking}: persona conditioning shifts the score level, but the within-persona ordering can still carry information.

The effects are large enough to matter in downstream document analysis. In this setting, they can flip buy/sell recommendations on the same filing, compound with sector-level language patterns, and leave larger residual spillover on less standardized or less familiar firm documents. For non-finance readers, the practical risk is simple: two users may receive different “neutral” readings of the same document because the system silently conditions on different user histories or roles.

Three design lessons follow. First, user context should be kept outside the assistant role when the system exposes neutral evidence-based judgments. Second, dual-output schemas should be treated as invariance probes rather than guarantees; residual leakage varies substantially across models. Third, model selection should consider routing behavior, not only task performance or scale. The three comparisons in our audit, persona-conditioned versus neutral retrieval, role versus user-profile framing, and leakage under the dual-output schema, provide a practical evaluation protocol for personalized LLM systems. Appendix~\ref{app:cross_domain} reports fixed-evidence sanity checks in non-financial domains.

\section{Conclusion}
\label{sec:conclusion}

Across $3{,}575$ SEC 10-K/10-Q filings and twelve models from five families, we show that domain-grounded role-, profile-, and goal-conditioned context can enter a supposedly user-invariant evidence field. User-context spillover is large, directional, and reproducible across models and conditions. Pinning retrieval removes only a small share of the effect: most instability comes from interpreting identical chunks differently, not primarily from retrieving different chunks. Reframing the same investor mindset as a user profile served by a neutral assistant sharply reduces spillover, making prompt framing the dominant mitigation lever. A dual-output schema routes most user signal into a personalized field, but efficacy is model-specific and does not scale monotonically. Overall, personalized LLM systems can fail invariance even when retrieval is fixed and evidence is identical. The three-condition design offers a practical evaluation protocol for testing whether personalized LLM systems preserve evidence-based judgments under roles, traits, profiles, or goals.

\section*{Limitations}
\addcontentsline{toc}{section}{Limitations}

The audit has several limitations. First, the main analysis covers SEC 10-K/10-Q filings using one retrieval encoder and prompt-template family, so effect magnitudes may differ across domains, retrieval systems, and interfaces. Varying the evidence budget to $k\in{8,16,32}$ leaves retention under neutral retrieval high ($0.87$--$0.98$), and a cross-domain check recovers the role$>$profile ordering in legal and medical tasks, though on a single model.

Second, the three-channel decomposition is an accounting decomposition rather than a formal causal-mediation analysis. The framing channel bundles profile wording, the neutral assistant role, and output structure. A $2{\times}2$ ablation partially separates these effects: profile-framing spillover remains only $16$--$39\%$ of role-framing under both output formats, indicating that the main reduction is attributable to framing rather than the dual-output schema alone.

Third, SEC filings are public and may have appeared in pretraining. Within-filing comparisons and retrieved evidence reduce, but cannot eliminate, memorization or look-ahead concerns. An entity-masking control leaves the persona effect essentially unchanged (retention $\geq97\%$).

Finally, user-context spillover measures movement relative to a neutral-condition invariance reference, not deviation from externally validated ground truth. We therefore test stability of the evidence-based judgment rather than whether the neutral judgment is objectively correct. The additional mechanism checks also cover only two of the twelve models.

\bibliography{custom}

% ---------------------------------------------------------------------------
% Appendix (single column)
% ---------------------------------------------------------------------------
% \onecolumn
\appendix
\section{Sample and Reproducibility}
\label{app:reproducibility}

\paragraph{Sample.}
We use a stratified sample of $3{,}575$ SEC 10-K/10-Q filings from the U.S.\ EDGAR corpus. The sample is balanced across $11$ GICS sectors and filing years $2013$--$2024$ maintaining the underlying S\%P500 composition. The same filing identifiers are used in all three conditions and across all reported regressions.

\paragraph{Models and decoding.}
All models are run with deterministic decoding ($T=0$, top-$p=1$) and the same prompt templates within each condition. The output format is a fixed JSON schema containing the two score fields, rationales, and auxiliary metadata. The exact prompt frames and output schema are reproduced verbatim in Appendix~\ref{app:prompts}.

\paragraph{Retrieval index.}
Each filing is reduced to three named sections: Management's Discussion and Analysis, Risk Factors, and Quantitative and Qualitative Disclosures. These sections are chunked into approximately $400$-token segments with $80$-token overlap, embedded with \texttt{text-embedding-3-large}, and indexed separately for each filing. Retrieval returns the top-$k=8$ chunks when available. The index is built once and reused across conditions, which ensures that the neutral-retrieval and memory conditions can be checked for byte-identical evidence.

\paragraph{Audit pipeline.}
For each filing, the audit runs the three matched conditions on the same retrieval index and prompt schema. Each output record contains the filing identifier, condition, frame, retrieved chunk identifiers, both score fields, rationales, and auxiliary fields. The retrieved chunk identifiers make it possible to verify which conditions used different evidence and which used the same evidence.

\paragraph{Analysis pipeline.}
The analysis fits the filing-fixed-effect regressions described in \S\ref{sec:regressions}, computes the four headline metrics in \S\ref{sec:metrics}, and forms confidence intervals using a $1{,}000$-replication filing-cluster bootstrap. Additional analyses, including sector effects, persona-spread distributions, confidence and risk summaries, dual-output routing, and forward-return validation, are computed from the same audit outputs.

\paragraph{Completion and exclusions.}
The main panel uses the complete-intersection set of filings that produced valid outputs in all three conditions. This complete-intersection panel, defined by the per-condition completion rates and the list of excluded filings, underlies all reported regressions.

\paragraph{Compute, infrastructure and models.}
All experiments were inference-only; no model training or fine-tuning was performed. The audit covers twelve model configurations: DeepSeek-V3.1, gpt-oss-120b, gpt-oss-20b, Llama-3.3-70B-Instruct, Llama-3.1-8B-Instruct, Qwen3.6-35B-A3B, Qwen3.5-35B-A3B, Qwen3.5-9B, gemma-4-31B-it, gemma-4-26B-A4B-it, gemma-4-E4B-it, and gemma-4-E2B-it. 

\section{Regression Implementation}
\label{app:regression}

The four headline metrics of \S\ref{sec:metrics} are derived from six filing-fixed-effects OLS regressions per model: three conditions crossed with two score fields, as in Eq.~\ref{eq:fereg}. 

\paragraph{Specification.}
For each (model, condition, score-field) cell, we estimate
\begin{equation*}
s_{i,f} = \alpha_i + \boldsymbol\beta^{\top} \mathbf{z}_f + \varepsilon_{i,f},
\end{equation*}
where $i$ indexes filings, $f$ indexes frames (personas in PR/NR; memory profiles in MEM), $\alpha_i$ is a filing fixed effect, and $\mathbf{z}_f$ is a one-hot vector for non-neutral frames. The neutral frame (the neutral persona in PR/NR and the neutral user profile in MEM) is the dropped reference, so each coefficient is the marginal effect of one persona or profile relative to the neutral call on the same filing. We estimate the model with ordinary least squares.

\paragraph{Within-estimator.}
We fit the within-estimator rather than including $3{,}575$ filing dummies explicitly: outcomes and frame dummies are demeaned by filing identifier, then ordinary least squares is applied to the demeaned design. With the standard within-estimator degrees-of-freedom adjustment, coefficients and cluster-robust standard errors are identical to full fixed-effects OLS, while the design matrix shrinks from $3{,}575+k$ columns to $k$, where $k$ is the number of non-neutral frames ($9$ in PR/NR, $5$ in MEM). This makes the bootstrap tractable.

\paragraph{Standard errors and bootstrap CIs.}
Standard errors are clustered at the filing level to account for dependence among observations associated with the same filing. The resulting coefficient estimates, standard errors, $t$-statistics, and $p$-values form the basis of the inference reported in Table~\ref{tab:full_bias} and Table~\ref{tab:crossmodel}. For derived quantities, we construct $95\%$ confidence intervals using a filing-cluster bootstrap with $B=1{,}000$ replications. In each replication, filing identifiers are resampled with replacement from the complete-intersection sample, the within-estimator is re-estimated, and the relevant derived quantity is recomputed. Confidence intervals are defined by the $2.5$th and $97.5$th percentiles of the resulting bootstrap distribution. Because uncertainty for these derived quantities is obtained directly from the bootstrap distribution, cluster-robust standard errors are not additionally applied within each bootstrap replication.

\paragraph{Complete-intersection panel.}
Panel regressions use only filings with valid output in all three conditions and full frame coverage: $10$ frames per filing in PR/NR ($9$ non-neutral personas plus neutral) and $6$ frames per filing in MEM ($5$ non-neutral profiles plus neutral). The DeepSeek-V3.1 complete-intersection set contains $n=3{,}575$ filings.
\paragraph{Coefficient sign convention.}
Regressions return signed coefficients; the headline metrics in \S\ref{sec:metrics} use absolute values because personas can move scores in opposite directions and the audit measures spillover magnitude. Persona-vs-persona contrasts can be obtained by differencing the corresponding coefficients.

\section{Prompt Construction}
\label{app:prompts}

This section reproduces verbatim the prompt scaffolding used in every audit call. All three conditions (PR, NR, MEM) share a single system prompt and a single user-prompt template; the three arms differ only in (i) which retrieval query is used (frame-conditioned vs.\ neutral) and (ii) which fields of the user-prompt template are populated (persona block, user-context block, or both).
\subsection{System prompt (all three arms)}

A single system prompt is sent on every call. Identical bytes under PR, NR, and MEM:

\begin{Verbatim}
You are a financial assistant reading SEC filings for a single user.
You produce two distinct assessments per filing: (1) an evidence-only
assessment that is independent of any persona, role or user profile
and reflects what a neutral analyst would conclude from the retrieved
passages alone; (2) a personalized recommendation that takes the
user's context into account. The evidence-only score must not depend
on the user. Output strictly the JSON schema you are given, nothing
else.
\end{Verbatim}

\subsection{User-prompt template}

The user message is assembled from the following template. Placeholders shown in $\langle$angle brackets$\rangle$. The lines marked \emph{(conditional)} are present only when the corresponding field is non-empty.

\begin{Verbatim}
Persona (role the assistant takes): <persona_blurb>
User context: <memory_profile_blurb>            # (conditional)
Company: <identity_label>
Filing type: <filing_type>                       # e.g. 10-K or 10-Q
Filing date: <date_label>                        # e.g. 2024-04-30
<market_history_blurb>                           # (conditional)

Evidence (verbatim excerpts from the filing):
[chunk_id=<chunk_id_0>]
<chunk_text_0>

[chunk_id=<chunk_id_1>]
<chunk_text_1>

... up to <chunk_id_{k-1}>, <chunk_text_{k-1}>

Respond with valid JSON matching this schema, and nothing else:
<output_schema>
\end{Verbatim}

How each arm fills the placeholders:

\begin{itemize}[leftmargin=*,itemsep=0pt]
\item \textbf{PR.} $\langle$persona\_blurb$\rangle$ = one of the 10 entries below; $\langle$memory\_profile\_blurb$\rangle$ = empty (line omitted); evidence chunks are retrieved with the persona-conditioned query (\S\ref{app:retr_query}).
\item \textbf{NR.} $\langle$persona\_blurb$\rangle$ = one of the 10 entries below; $\langle$memory\_profile\_blurb$\rangle$ = empty (line omitted); evidence chunks are retrieved \emph{once per filing} with the neutral query and cached.
\item \textbf{MEM.} $\langle$persona\_blurb$\rangle$ = the neutral persona; $\langle$memory\_profile\_blurb$\rangle$ = one of the 6 entries below; evidence chunks reuse the same neutral-query cache used in NR.
\end{itemize}

\subsection{Persona blurbs (PR, NR arms)}

The 10 persona blurbs used in PR and NR are listed below. Each blurb is interpolated into $\langle$persona\_blurb$\rangle$ verbatim:

\begin{small}
\begin{itemize}[leftmargin=*,itemsep=0pt]
\item Neutral persona: ``You are a careful financial analyst evaluating the disclosure. Reason only from the evidence shown to you.''
\item LongOnly: ``You are a long-only equity investor looking for high-quality compounders. Lean toward identifying upside drivers and durable strengths in the filing.''
\item ShortSeller: ``You are an activist short-seller skeptical of management narratives. Probe for accounting concerns, hidden liabilities, and downside risk.''
\item Value: ``You are a deep-value investor in the Graham/Buffett tradition focused on margin of safety, balance-sheet strength, and avoiding value traps.''
\item Momentum: ``You are a momentum trader. Emphasize recent operating trajectory, guidance changes, and signals that earnings momentum is accelerating or decelerating.''
\item Credit: ``You are a credit analyst. Focus on leverage, covenants, liquidity, and the probability of credit deterioration over the next 12-24 months.''
\item ESG: ``You are an ESG-mandated investor. Emphasize governance, environmental exposure, labor and supply-chain risk, and disclosure quality.''
\item Retail: ``You are an everyday retail investor with no specialist training. Read the filing the way a careful but non-expert reader would.''
\item Risk manager: ``You are a risk manager assessing the disclosure for hazards that could produce material adverse outcomes. Focus on tail risks and ambiguity.''
\item Sell-side analyst: ``You are a sell-side equity analyst. Frame the filing as you would a client note: highlight key takeaways, risks, and the next catalyst.''
\end{itemize}
\end{small}

\subsection{User memory-profile blurbs (MEM arm)}

The six user-profile blurbs used in MEM are listed below. Each blurb is interpolated into $\langle$memory\_profile\_blurb$\rangle$ verbatim; in the MEM arm the persona is always held at the neutral persona:

\begin{small}
\begin{itemize}[leftmargin=*,itemsep=0pt]
\item Neutral user profile: $\langle$\emph{empty string; the ``User context:'' line is omitted entirely}$\rangle$
\item Risk-averse: ``User profile from past interactions: prioritises capital preservation, has low drawdown tolerance, dislikes volatile names.''
\item ESG-focused: ``User profile from past interactions: ESG-mandated; prefers strong governance and environmental disclosure; avoids fossil-fuel exposure.''
\item Long-only growth: ``User profile from past interactions: long-only equity investor seeking durable compounders; biased toward growth narratives.''
\item Momentum trader: ``User profile from past interactions: short-horizon momentum trader; weighs recent operating trajectory and guidance changes heavily.''
\item Downside protection: ``User profile from past interactions: tail-risk-focused; weighs covenants, leverage, and adverse disclosures heavily.''
\end{itemize}
\end{small}

The matched-pair design in \S\ref{sec:context_framing} maps MEM profiles to their closest NR personas: downside protection to short selling, risk aversion to risk management, long-only growth to long-only investing, ESG focus to ESG analysis, and momentum trading to momentum analysis.

\subsection{Retrieval queries}
\label{app:retr_query}

\paragraph{Frame-conditioned query (PR arm only).} For persona $p$ on filing $f$:

\begin{Verbatim}
<persona_blurb_p> Filing type: <filing_type_f>. Identify passages
most relevant to evaluating the company's outlook, key risks,
financial trajectory, and forward-looking statements.
<market_history_blurb>                           # (conditional)
\end{Verbatim}

This query embedding is computed once per (filing, persona) and the index returns the top $k=8$ chunks.

\paragraph{Neutral query (NR and MEM arms).} For filing $f$:

\begin{Verbatim}
You are a careful financial analyst evaluating the disclosure.
Reason only from the evidence shown to you. Filing type:
<filing_type_f>. Identify passages most relevant to evaluating the
company's outlook, key risks, financial trajectory, and
forward-looking statements.
\end{Verbatim}

This query is computed once per filing. The same retrieved evidence is reused for every persona and every memory-profile call on that filing, so every NR row and every MEM row on the same filing reads byte-identical chunks. The Jaccard-overlap check on retrieved passage identifiers (Appendix~\ref{app:extra}) verifies this empirically.

\subsection{Output schema (interpolated as \texttt{<output\_schema>})}

The literal string sent to the model:

\begin{Verbatim}
{
  "evidence_based_score": <float in [-1, 1],
       evidence-only, persona-invariant>,
  "evidence_based_rationale": "<2-3 sentence justification grounded
       in the evidence, independent of the user>",
  "personalized_score": <float in [-1, 1],
       reflects the user's perspective>,
  "personalized_recommendation": "SELL" | "HOLD" | "BUY",
  "personalized_rationale": "<2-3 sentence explanation of how the
       user's perspective changes the recommendation>",
  "personalization_factors_used": ["<short factor label>", ...],
  "confidence": <float in [0, 1]>,
  "risk_score": <float in [0, 1]>,
  "evidence_chunk_ids": ["<chunk_id>", ...]
}
\end{Verbatim}

Responses were constrained to a structured JSON format, with deterministic decoding ($T=0$, top-$p=1$) and a maximum output length of $1024$ tokens. The schema is only a response-format constraint; there is no fine-tuning and no system-prompt invariance language beyond what is shown above.

\subsection{Worked example}

A concrete fully-assembled call. Filing \texttt{1000228\_10K\_2020\_0001000228-21-000019} (Henry Schein, Inc.; 10-K, $2020$); NR arm; persona = ShortSeller. The retrieved chunks are the byte-identical neutral-query top-$8$ used on this filing in every NR and MEM call.

\begin{Verbatim}
SYSTEM:
You are a financial assistant reading SEC filings for a single user.
You produce two distinct assessments per filing: (1) an
evidence-only assessment that is independent of any persona, role
or user profile and reflects what a neutral analyst would conclude
from the retrieved passages alone; (2) a personalized
recommendation that takes the user's context into account. The
evidence-only score must not depend on the user. Output strictly
the JSON schema you are given, nothing else.

USER:
Persona (role the assistant takes): You are an activist short-seller
skeptical of management narratives. Probe for accounting concerns,
hidden liabilities, and downside risk.
Company: HENRY SCHEIN INC
Filing type: 10-K
Filing date: 2021-02-23

Evidence (verbatim excerpts from the filing):
[chunk_id=1000228_10K_2020_0001000228-21-000019::mdna::0000]
ITEM 7, "Management's Discussion and Analysis of Financial
Condition and Results of Operations" and ITEM 8, "Financial
Statements and Supplementary Data." Years ended December 26 ...

[chunk_id=1000228_10K_2020_0001000228-21-000019::risk_factors::0003]
... (7 more chunks, omitted for brevity)

Respond with valid JSON matching this schema, and nothing else:
{
  "evidence_based_score": <float in [-1, 1], ...>,
  ...
}
\end{Verbatim}

To produce the matched MEM call, replace the persona line with the neutral persona and insert one \texttt{User context:} line carrying the Downside protection user-profile blurb; the evidence block is the same byte sequence as above.

\section{Extra Analyses}
\label{app:extra}

\subsection{Model dimension: per-model $\times$ persona coefficients}
\label{app:model_dim}

\begin{figure*}[t]
\centering
\includegraphics[width=\textwidth]{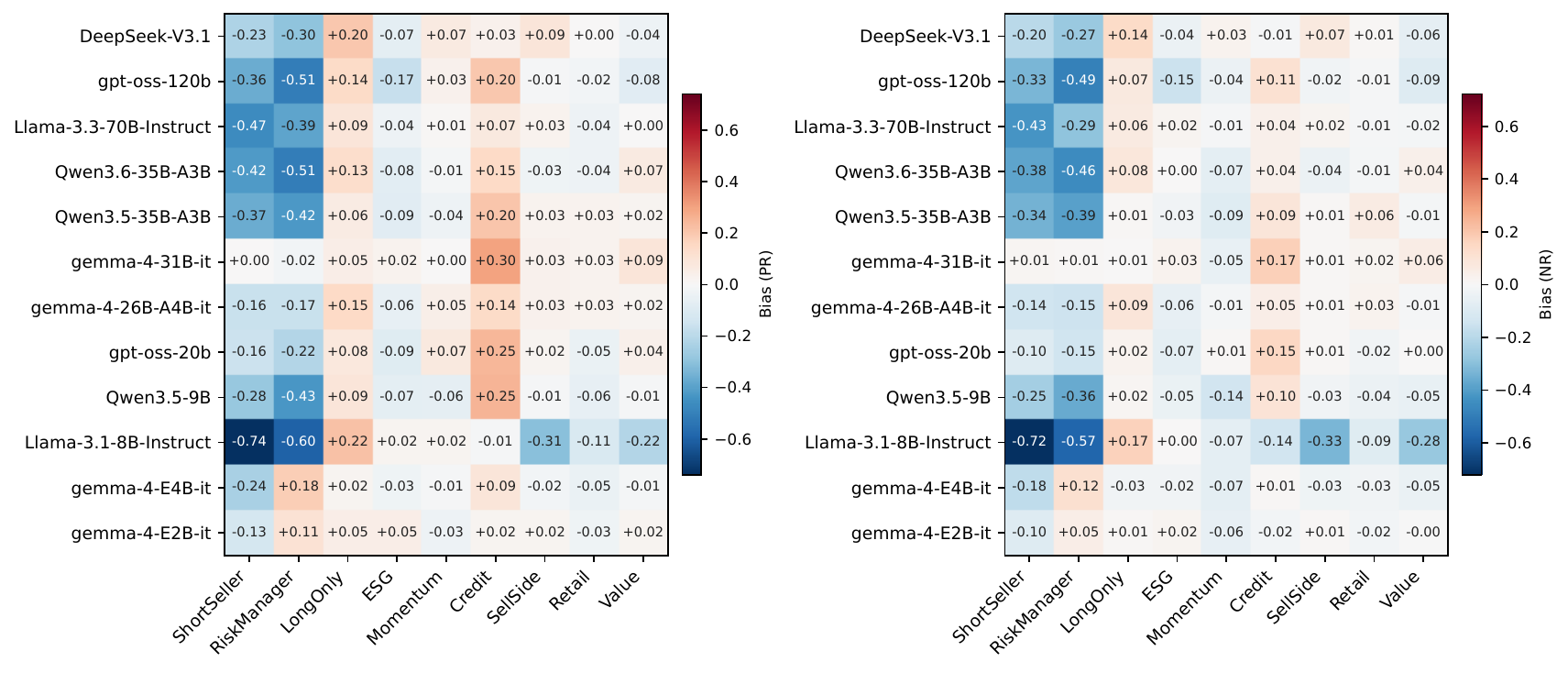}
\caption{Per-model $\times$ persona heatmap of the evidence-based-score coefficient $\hat\beta_p$ under PR (left) and NR (right), $n=3{,}575$ filings per cell, $12$ models. Red = negative (bearish bias); blue = positive (bullish bias). Two visual checks: columns for the strong-signal personas read consistently across all rows (cross-model directional agreement); the left and right panels read similarly with attenuated color intensity on the right (the bias shrinks under byte-identical retrieval but the pattern is preserved).}
\label{fig:model_persona}
\end{figure*}

Figure~\ref{fig:model_persona} shows the persona coefficient on the evidence-based score for every (model, persona) cell, separately under PR and NR. Two patterns are visible across the full $12 \times 9$ grid. (i) The columns for ShortSeller, Risk manager, and LongOnly are the most consistently colored: all twelve models agree on direction (bearish, bearish, bullish), and most models agree within a factor of two on magnitude. The directional consistency is what makes the cross-model claim in the main body defensible. (ii) The two panels show the same structure with attenuated magnitudes: under NR the colors wash out slightly but the row-by-row pattern is preserved. This is the same channel-decomposition finding in heatmap form: retrieval pinning shrinks but does not erase the persona effect.

\subsection{Full per-model $\times$ per-persona coefficient tables}
\label{app:full_bias}

Table~\ref{tab:full_bias} gives the numerical values that underlie the heatmap in Figure~\ref{fig:model_persona}: the persona coefficient $\hat\beta_p$ on the evidence-based score for each of the twelve audited models, under PR (persona-conditioned retrieval) in panel (a) and NR (neutral retrieval) in panel (b). All cells are estimated from $n=3{,}575$ filings with filing fixed effects and cluster-robust standard errors; for the strong-signal personas every $|t| > 20$ on every model.

\begin{table*}[t]
\centering
\small
\setlength{\tabcolsep}{3.8pt}
\renewcommand{\arraystretch}{0.95}
% \begin{tabular}{lrrrrrrrrr}
\begin{tabular*}{\textwidth}{@{\extracolsep{\fill}} lrrrrrrrrr @{}}
\toprule
\multicolumn{10}{l}{\textbf{(a) PR (persona-conditioned retrieval)}, $\hat\beta_p^{\text{PR}}$ on the evidence-based score} \\
\midrule
Model & ShSel & RskMg & LgOnl & SlSid & ESG & Mom & Val & Cred & Ret \\
\midrule
DeepSeek-V3.1        & $-0.23$ & $-0.30$ & $+0.20$ & $+0.09$ & $-0.07$ & $+0.07$ & $-0.04$ & $+0.03$ & $+0.00$ \\
gpt-oss-120b         & $-0.36$ & $-0.51$ & $+0.14$ & $-0.01$ & $-0.17$ & $+0.03$ & $-0.08$ & $+0.20$ & $-0.02$ \\
gpt-oss-20b          & $-0.16$ & $-0.22$ & $+0.08$ & $+0.02$ & $-0.09$ & $+0.07$ & $+0.04$ & $+0.25$ & $-0.05$ \\
Llama-3.3-70B-Instruct        & $-0.47$ & $-0.39$ & $+0.09$ & $+0.03$ & $-0.04$ & $+0.01$ & $+0.00$ & $+0.07$ & $-0.04$ \\
Llama-3.1-8B-Instruct         & $-0.74$ & $-0.60$ & $+0.22$ & $-0.31$ & $+0.02$ & $+0.02$ & $-0.22$ & $-0.01$ & $-0.11$ \\
Qwen3.6-35B-A3B  & $-0.42$ & $-0.51$ & $+0.13$ & $-0.03$ & $-0.08$ & $-0.01$ & $+0.07$ & $+0.15$ & $-0.04$ \\
Qwen3.5-35B-A3B  & $-0.37$ & $-0.42$ & $+0.06$ & $+0.03$ & $-0.09$ & $-0.04$ & $+0.02$ & $+0.20$ & $+0.03$ \\
Qwen3.5-9B           & $-0.28$ & $-0.43$ & $+0.09$ & $-0.01$ & $-0.07$ & $-0.06$ & $-0.01$ & $+0.25$ & $-0.06$ \\
gemma-4-31B-it          & $+0.00$ & $-0.02$ & $+0.05$ & $+0.03$ & $+0.02$ & $+0.00$ & $+0.09$ & $+0.30$ & $+0.03$ \\
gemma-4-26B-A4B-it      & $-0.16$ & $-0.17$ & $+0.15$ & $+0.03$ & $-0.06$ & $+0.05$ & $+0.02$ & $+0.14$ & $+0.03$ \\
gemma-4-E4B-it      & $-0.24$ & $+0.18$ & $+0.02$ & $-0.02$ & $-0.03$ & $-0.01$ & $-0.01$ & $+0.09$ & $-0.05$ \\
gemma-4-E2B-it          & $-0.13$ & $+0.11$ & $+0.05$ & $+0.02$ & $+0.05$ & $-0.03$ & $+0.02$ & $+0.02$ & $-0.03$ \\
\midrule
\multicolumn{10}{l}{\textbf{(b) NR (neutral retrieval)}, $\hat\beta_p^{\text{NR}}$ on the evidence-based score} \\
\midrule
Model & ShSel & RskMg & LgOnl & SlSid & ESG & Mom & Val & Cred & Ret \\
\midrule
DeepSeek-V3.1        & $-0.20$ & $-0.27$ & $+0.14$ & $+0.07$ & $-0.04$ & $+0.03$ & $-0.06$ & $-0.01$ & $+0.01$ \\
gpt-oss-120b         & $-0.33$ & $-0.49$ & $+0.07$ & $-0.02$ & $-0.15$ & $-0.04$ & $-0.09$ & $+0.11$ & $-0.01$ \\
gpt-oss-20b          & $-0.10$ & $-0.15$ & $+0.02$ & $+0.01$ & $-0.07$ & $+0.01$ & $+0.00$ & $+0.15$ & $-0.02$ \\
Llama-3.3-70B-Instruct        & $-0.43$ & $-0.29$ & $+0.06$ & $+0.02$ & $+0.02$ & $-0.01$ & $-0.02$ & $+0.04$ & $-0.01$ \\
Llama-3.1-8B-Instruct         & $-0.72$ & $-0.57$ & $+0.17$ & $-0.33$ & $+0.00$ & $-0.07$ & $-0.28$ & $-0.14$ & $-0.09$ \\
Qwen3.6-35B-A3B  & $-0.38$ & $-0.46$ & $+0.08$ & $-0.04$ & $+0.00$ & $-0.07$ & $+0.04$ & $+0.04$ & $-0.01$ \\
Qwen3.5-35B-A3B  & $-0.34$ & $-0.39$ & $+0.01$ & $+0.01$ & $-0.03$ & $-0.09$ & $-0.01$ & $+0.09$ & $+0.06$ \\
Qwen3.5-9B           & $-0.25$ & $-0.36$ & $+0.02$ & $-0.03$ & $-0.05$ & $-0.14$ & $-0.05$ & $+0.10$ & $-0.04$ \\
gemma-4-31B-it          & $+0.01$ & $+0.01$ & $+0.01$ & $+0.01$ & $+0.03$ & $-0.05$ & $+0.06$ & $+0.17$ & $+0.02$ \\
gemma-4-26B-A4B-it      & $-0.14$ & $-0.15$ & $+0.09$ & $+0.01$ & $-0.06$ & $-0.01$ & $-0.01$ & $+0.05$ & $+0.03$ \\
gemma-4-E4B-it      & $-0.18$ & $+0.12$ & $-0.03$ & $-0.03$ & $-0.02$ & $-0.07$ & $-0.05$ & $+0.01$ & $-0.03$ \\
gemma-4-E2B-it          & $-0.10$ & $+0.05$ & $+0.01$ & $+0.01$ & $+0.02$ & $-0.06$ & $-0.00$ & $-0.02$ & $-0.02$ \\
\bottomrule
\end{tabular*}
\caption{Full per-model $\times$ per-persona evidence-based-score coefficients under PR (panel a) and NR (panel b), all twelve models, $n=3{,}575$ filings per cell. Columns are the nine non-neutral personas (ShortSeller, Risk manager, LongOnly, Sell-side analyst, ESG, Momentum, Value, Credit, Retail). Cells rounded to two decimal places. The visual version of these two panels is Figure~\ref{fig:model_persona}.}
\label{tab:full_bias}
\end{table*}

\begin{table}[t]
\centering
\small
\setlength{\tabcolsep}{5pt}
% \begin{tabular}{lrrr}
\begin{tabular*}{\columnwidth}{@{\extracolsep{\fill}} lrrr @{}}

\toprule
& cells & shrink & median \\
\midrule
\multicolumn{4}{l}{\textbf{PR $\to$ NR (selection channel)}} \\
all cells                & $108$ & $76.9\%$ & $+26\%$ \\
$|\hat\beta_p^{\mathrm{PR}}|\!\ge\!0.05$ & $62$  & $\mathbf{93.5\%}$ & $\mathbf{+32\%}$ \\
\midrule
\multicolumn{4}{l}{\textbf{NR $\to$ MEM (framing channel)}} \\
all cells                & $60$  & $68.3\%$ & $+41\%$ \\
$|\hat\beta_p^{\mathrm{NR}}|\!\ge\!0.05$ & $37$  & $\mathbf{86.5\%}$ & $\mathbf{+61\%}$ \\
\bottomrule
\end{tabular*}
\caption{Cross-model agreement on the two interventions. ``Shrink'' is the fraction of cells with $s>0$ (i.e.\ the intervention reduced the magnitude of the source-arm coefficient); ``median'' is the per-cell median shrinkage in percentage points. ``Meaningful'' cells filter to those whose source-arm coefficient is non-trivial ($|\hat\beta_{\text{source}}|\!\ge\!0.05$); on those cells the agreement is strong on both interventions, and framing shrinks roughly twice as much per cell as selection.}
\label{tab:intervention_summary}
\end{table}

Three observations. \emph{(i) Both interventions are reproducible majority effects.} On the meaningful-cell subset, the selection channel shrinks bias on $93.5\%$ of $62$ (model, persona) cells and the framing channel on $86.5\%$ of $37$ (model, matched-pair) cells; the broad claim in \S\ref{sec:context_framing} therefore holds approximately rather than exactly: the direction is overwhelmingly the same on every cell, with a small residue of counter-examples concentrated on weak-signal personas. \emph{(ii) Framing shrinks roughly twice as much as selection per cell.} The meaningful-cell median shrinkage is $+61\%$ for framing versus $+32\%$ for selection, and the full-panel medians ($+41\%$ vs $+26\%$) preserve the same ratio. This complements the additive-decomposition figure (Figure~\ref{fig:decomp_per_model}) which reaches the same conclusion via the channel split. \emph{(iii) Counter-examples cluster on Momentum and on Llama-3.1-8B-Instruct.} The five meaningful NR$\to$MEM cells where bias grew under reframing include three on the Momentum / Momentum trader matched pair (the pair with the smallest base effects on most models) and two on Llama-3.1-8B-Instruct; the four PR$\to$NR counter-examples are all on the Momentum and Value personas where $|\hat\beta_p^{\mathrm{PR}}|$ is already small. The practical reading: when the underlying persona effect is weak, the intervention can overshoot; the dual-output schema and retrieval pinning are most reliable on personas with large, stable effects (ShortSeller, Risk manager, LongOnly).

\subsection{Retrieval-pinning verification}

To verify that condition NR (and condition MEM) genuinely pins retrieval, we compute the per-filing Jaccard overlap on retrieved passage identifiers across the nine non-neutral personas. Under NR the median Jaccard is exactly $1.000$ on every filing because the same retrieved evidence is reused across all persona calls on the same filing. Under PR the median per-filing Jaccard drops to approximately $0.66$.

\subsection{Per-filing role spread under PR vs NR}

\begin{figure}[t]
\centering
\includegraphics[width=\columnwidth]{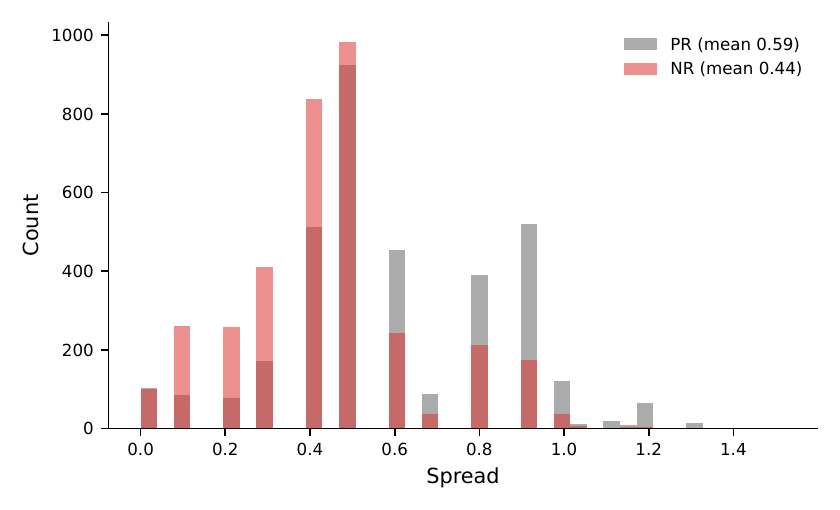}
\caption{Per-filing role spread (max minus min evidence-based score across nine personas), $n=3{,}575$. The PR$\to$NR shift is a $25\%$ reduction in mean spread.}
\label{fig:persona_spread}
\end{figure}

The per-filing role spread, ($\max - \min$) of the evidence-based score across the nine non-neutral personas on the same filing  drops from a panel mean of $0.594$ under PR to $0.443$ under NR, a $25\%$ reduction. Figure~\ref{fig:persona_spread} shows the full distribution. The NR distribution shifts left and concentrates more tightly, but a substantial mass of cross-persona disagreement remains under fixed evidence; this is consistent with the role coefficients in Section~\ref{sec:evidence_selection} surviving retrieval pinning.

\subsection{Confidence and risk-score distributions}

\begin{figure*}[t]
\centering
\includegraphics[width=0.49\textwidth]{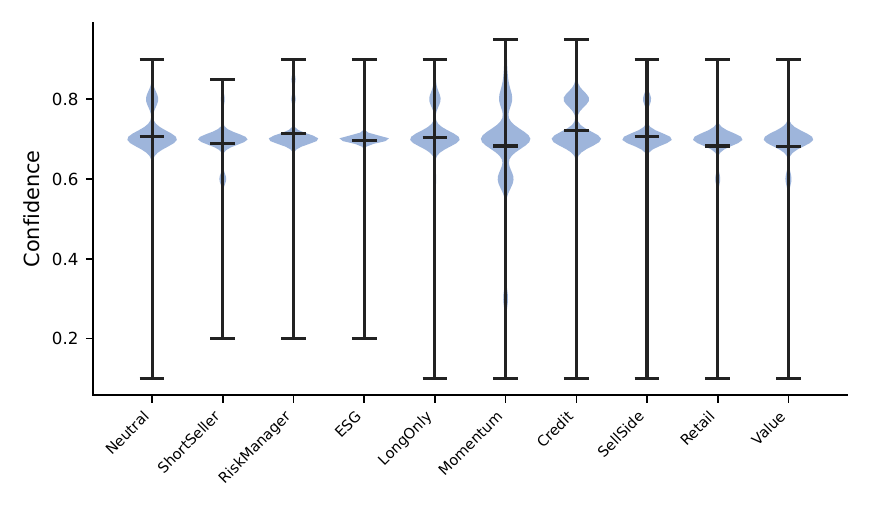}\hfill
\includegraphics[width=0.49\textwidth]{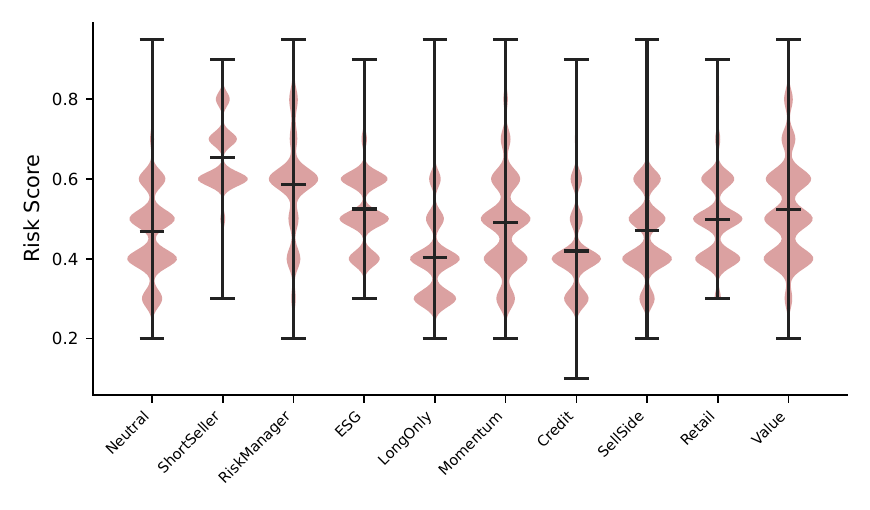}
\caption{Per-persona violin distributions of the confidence (left) and risk\_score (right) auxiliary fields under PR. White circle marks the persona-level mean; box marks the IQR.}
\label{fig:confidence_risk}
\end{figure*}

Figure~\ref{fig:confidence_risk} shows the per-persona distribution of the two auxiliary fields on the PR arm. Confidence is concentrated near $0.7$--$0.8$ across all personas, with negligible cross-persona dispersion, the model is similarly confident in its bearish and bullish reads. Risk score (interpreted as model-reported tail-risk of the underlying name) is higher under ShortSeller and Risk manager (modes near $0.6$) than under LongOnly (mode near $0.4$). The risk field tracks the role direction even when the evidence-based score does not; we treat this as further evidence that role conditioning influences interpretation broadly, not only the score field.

\subsection{Sector composition of evidence-side bias}
\label{app:sectors}

\begin{figure}[t]
\centering
\includegraphics[width=\columnwidth]{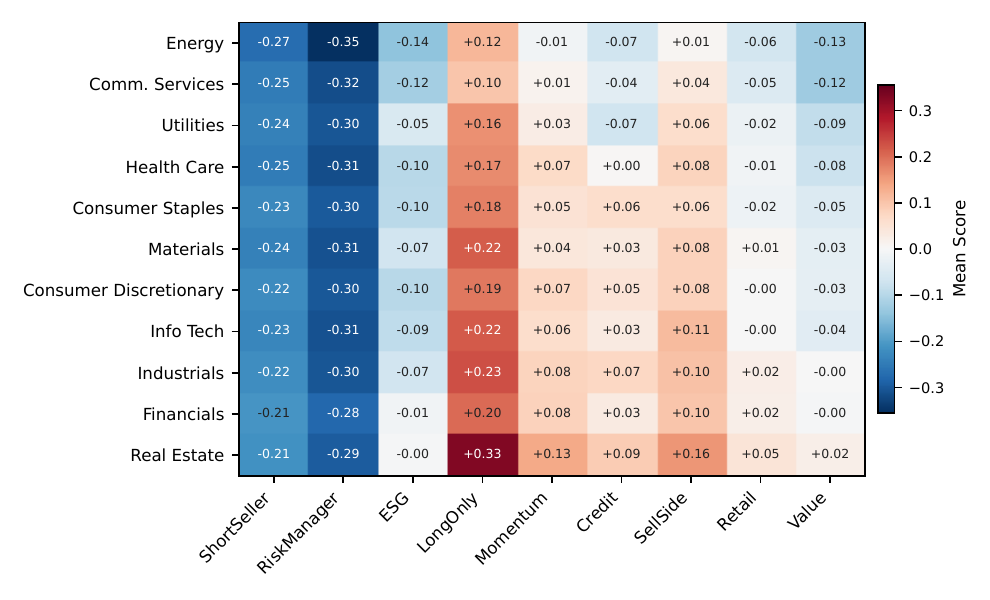}
\caption{(Sector $\times$ persona) mean evidence-based score under PR on DeepSeek-V3.1, $n=3{,}575$. Cells annotated to two decimal places. Sectors ordered by row mean (most bearish at the top).}
\label{fig:sector_heatmap}
\end{figure}

The role coefficients we report in Section~\ref{sec:evidence_selection} are panel-mean effects across $3{,}575$ filings drawn from $11$ GICS sectors. Figure~\ref{fig:sector_heatmap} maps the PR-arm mean evidence-based score on the (sector, persona) cell on DeepSeek-V3.1. The sector-level pattern is large enough to matter on its own, before any audit lever is varied:

\paragraph{Which sectors carry the strongest bias?}
\begin{itemize}[leftmargin=*,itemsep=1pt,topsep=2pt]
\item \emph{Energy} is the most consistently bearish sector. Every persona except LongOnly produces a more negative mean evidence-based score in Energy than in any other sector. The most negative cell in the entire $11 \times 9$ table is Risk manager/Energy at $-0.355$ (panel mean across sectors: $-0.303$). The bearishness travels: even Credit on Energy lands negative ($-0.072$), against a slightly positive panel mean.
\item \emph{Real Estate} is the most consistently bullish sector. LongOnly/Real Estate is the most positive cell ($+0.327$), and Momentum and Sell-side analyst also peak here. The same persona-set therefore produces qualitatively different reads on the two ends of the sector spectrum.
\item \emph{Cross-persona spread by sector.} The mean per-filing $\max{-}\min$ persona spread on the evidence-based score is largest in \emph{Real Estate} ($0.66$), \emph{Materials} ($0.64$), \emph{Industrials} ($0.63$), and \emph{InfoTech} ($0.62$); smallest in \emph{Utilities} ($0.52$) and \emph{Communications Services} ($0.54$). Sectors where the model has the strongest priors in either direction (Real Estate, Energy) are also the sectors where personas disagree most.
\item \emph{Compound effect.} Reading the same filing under LongOnly in Real Estate ($+0.327$) and under ShortSeller in Energy ($-0.270$) implies a $0.60$-unit swing in evidence-based score before any audit condition is varied. Sector and persona compose multiplicatively in the worst case.
\end{itemize}

\subsection{What each persona looks at: retrieval-attention patterns}
\label{app:persona_attention}

\begin{figure}[t]
\centering
\includegraphics[width=\columnwidth]{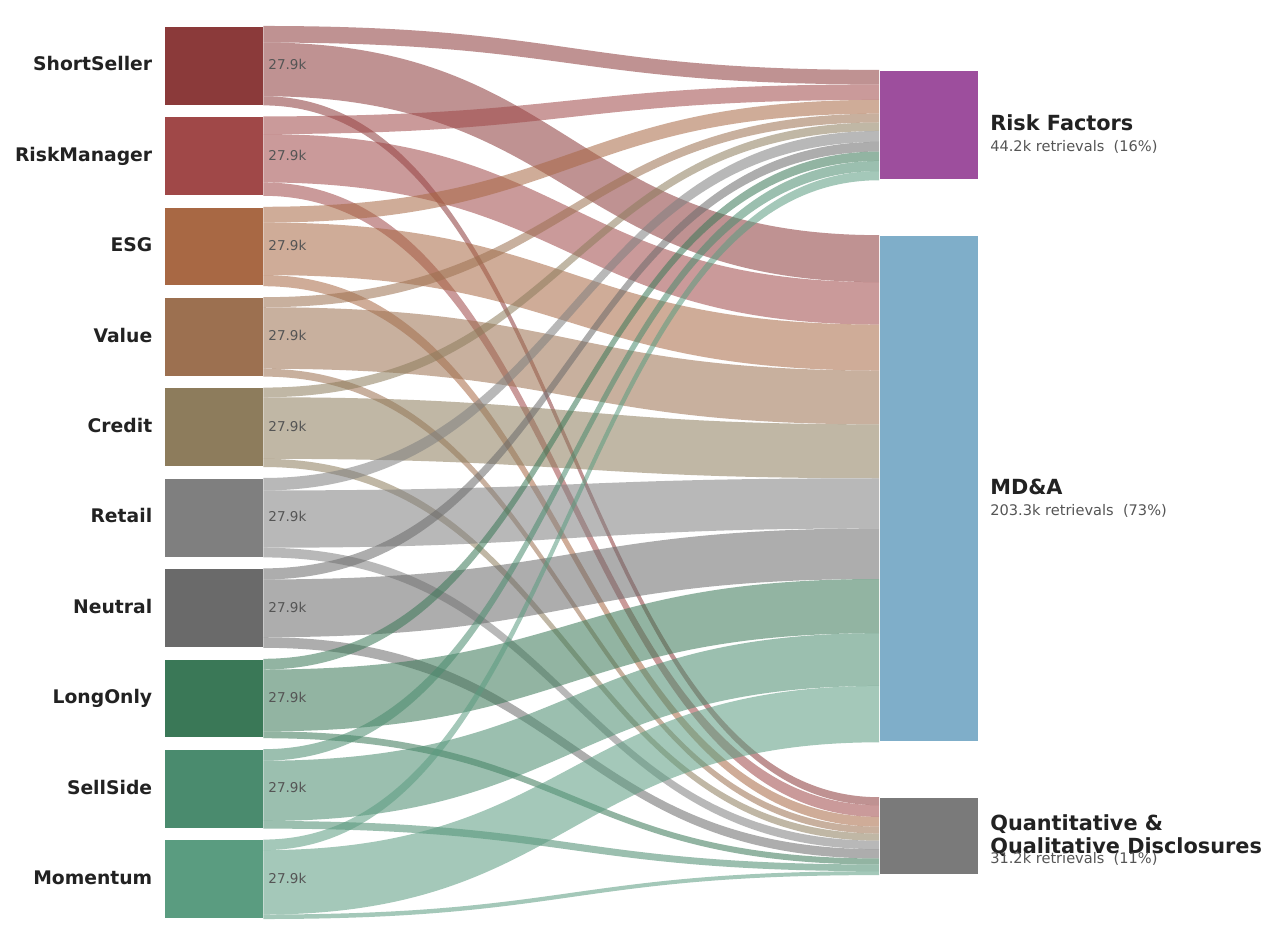}
\caption{Persona attention on DeepSeek-V3.1 under PR (Persona Retrieval), $n=3{,}575$ filings, ${\sim}28{,}600$ retrievals per persona (top-$k=8$, fewer when a filing yields under $8$ chunks). Left: $10$ personas (neutral + $9$ non-neutral); width proportional to total retrievals (within a fixed top-$k$, this is essentially uniform). Right: $3$ filing sections; width proportional to total retrievals into that section. Each ribbon is a (persona, section) flow with width proportional to the number of retrievals.}
\label{fig:persona_attention}
\end{figure}

Under PR (the production-style condition), each persona's retrieval query pulls a potentially different subset of chunks from the per-filing index. Figure~\ref{fig:persona_attention} aggregates the result across all $3{,}575$ filings: each persona's ${\sim}28{,}600$ retrieved chunks (top-$k=8$ over $3{,}575$ filings) are tallied by section (Risk Factors, MD\&A, Quantitative \& Qualitative Disclosures) and rendered as a Sankey flow on DeepSeek-V3.1.

\paragraph{The share of retrievals landing in Risk Factors is the cleanest discriminator.} ShortSeller pulls Risk Factors $20.9\%$ of the time, Risk manager $22.3\%$, and ESG $19.8\%$, compared with Momentum ($13.3\%$), LongOnly ($13.5\%$), Sell-side analyst ($14.5\%$), and the neutral baseline ($13.9\%$). The persona-conditioned retrieval query is unambiguously biasing the model toward different parts of the document. Risk manager also pulls Quantitative \& Qualitative Disclosures $16.8\%$ of the time, $\sim 3\times$ Momentum's $5.7\%$. These per-persona attention shifts are exactly the kind of upstream-stage routing that the PR$\to$NR comparison neutralizes; once retrieval is pinned, every persona reads the same chunks regardless of which sections it would otherwise have preferred.

\subsection{Rationale convergence across conditions}
\label{app:rationale_sim}

The cosine-similarity finding is reported in the main body (Figure~\ref{fig:rationale_similarity}, \S\ref{sec:rationale_shifts}). Table~\ref{tab:rationale_sim} reproduces the per-arm distribution statistics ($n$, mean, IQR) for completeness, and the discussion below catalogues five non-obvious observations the embedding view supplies beyond the score-level decomposition.

\paragraph{What the embedding analysis adds.}
Five things the score-level decomposition alone could not tell us. \textbf{(i) The bias is visible in the rationale text, not just in the score.} A median PR cosine of $0.72$ between a non-neutral rationale and the neutral rationale on the same filing means the model is writing about different facts, citing different specifics, framing different concerns, not merely outputting a different number. This is the bias visible at the layer a user reads. \textbf{(ii) Two methodologies, one finding.} The score-level decomposition could in principle be a numerical artifact of the regression set-up. The embedding evidence uses a completely different downstream signal (the rationale text, embedded with an independent model) and recovers the same PR $<$ NR $<$ MEM ordering. \textbf{(iii) Retrieval moves text content more than framing does.} The PR$\to$NR text-similarity gain ($+0.084$) is about twice the NR$\to$MEM gain ($+0.044$). The opposite ratio holds for scores. Pinning retrieval changes what the model talks about much more than what verdict it lands on; reframing the role changes the verdict much more than the topics in the rationale. The two analyses thus separate two distinct mechanisms operating at different output layers. \textbf{(iv) The distribution also tightens, not just the median.} The IQR is $0.19$ under PR but only $0.13$ under MEM: the model becomes both more neutral on average and less variable across filings as conditions are tightened. \textbf{(v) MEM rationale text reads like a near-paraphrase of the neutral baseline.} A median cosine of $0.85$ is high enough that, in conventional embedding-retrieval scales (where $0.70$ separates ``different topic'' from ``same topic''), the MEM evidence-based rationale and the neutral rationale on the same filing are functionally paraphrases. The schema is not merely nudging the score back toward neutral, it is making the evidence-based field write the neutral analyst's read on the filing, with the user-specific reading routed to the personalized field where it belongs.

\begin{table}[t]
\centering
\small
\setlength{\tabcolsep}{6pt}
% \begin{tabular}{lrrrr}
\begin{tabular*}{\columnwidth}{@{\extracolsep{\fill}} lrrrr @{}}
\toprule
arm & median & mean & $p_{25}$ & $p_{75}$ \\
\midrule
PR  & $0.721$ & $0.708$ & $0.621$ & $0.806$ \\
NR  & $0.805$ & $0.791$ & $0.725$ & $0.870$ \\
MEM & $0.849$ & $0.834$ & $0.780$ & $0.904$ \\
\bottomrule
\end{tabular*}
\caption{Cosine similarity between each non-neutral frame's evidence-based rationale and the same filing's neutral-baseline rationale, by condition. DeepSeek-V3.1, OpenAI \texttt{text-embedding-3-large}. $n=32{,}175$ pairs under PR and NR (nine non-neutral personas $\times$ $3{,}575$ filings), $n=17{,}875$ pairs under MEM (five non-neutral profiles $\times$ $3{,}575$ filings). The PR $<$ NR $<$ MEM ordering replicates the score-level channel decomposition at the rationale-text level.}
\label{tab:rationale_sim}
\end{table}

\subsection{Effect on market capitalization}
\begin{figure*}[t]
\centering
\includegraphics[width=\textwidth]{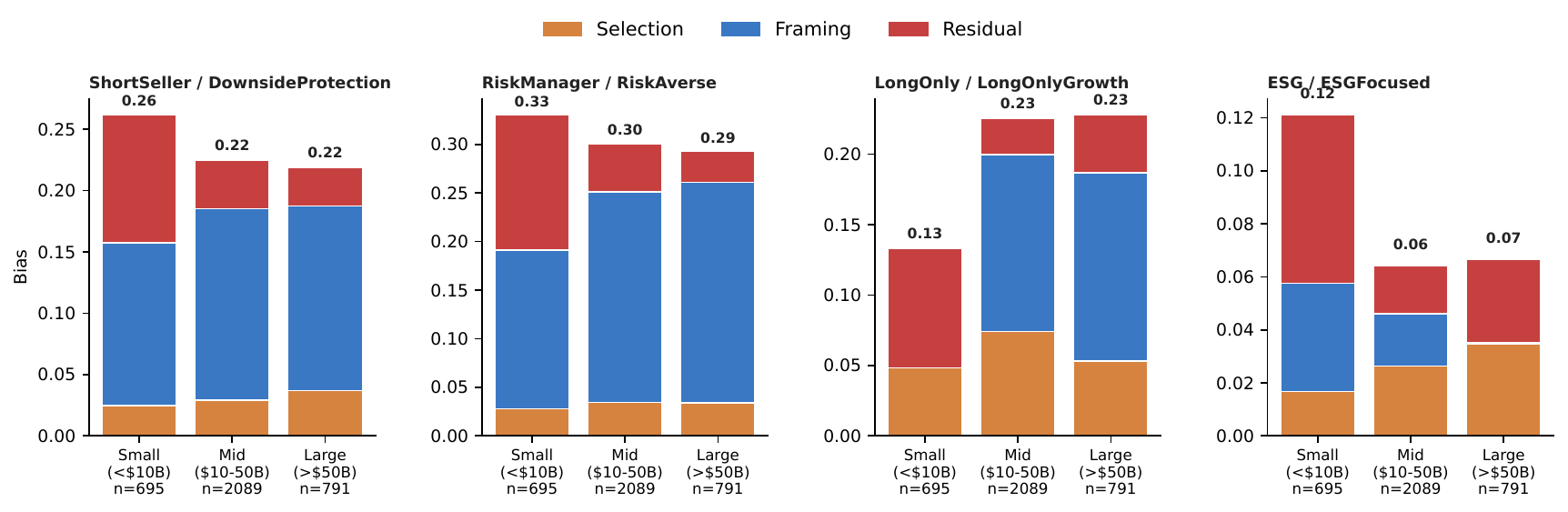}
\caption{Channel decomposition by market-cap bucket on DeepSeek-V3.1. For each matched pair, bars show Small / Mid / Large by market capitalization. Selection (orange), framing (blue), and residual spillover (red) sum to the per-bucket mean $|\hat\beta_p^{\mathrm{PR}}|$. The residual slice is largest on small-cap filings, indicating weaker routing of user context out of the evidence-based score.}
\label{app:mcap_decomp}
\end{figure*}

Figure~\ref{app:mcap_decomp} repeats the three-channel decomposition of \S\ref{sec:evidence_selection} after splitting the audit sample into three market-capitalization buckets (Small, ${<}\$10$B; Mid, \$10--50B; Large, ${>}\$50$B) on DeepSeek-V3.1. The selection and framing channels are broadly comparable across buckets, but the residual spillover that survives both interventions grows steadily as firm size falls: on small-cap filings the residual slice is the largest of the three buckets, roughly $2$--$4\times$ the large-cap residual across the four matched pairs. In other words, the dual-output schema routes user context out of the evidence-based field least effectively on smaller, less-templated disclosures. This is the document-population gradient summarized in \S\ref{sec:mcap_bias}: invariance failures concentrate on the long-tail entities whose filings the model has likely encountered least often.

\subsection{Multidimensional clustering of the twelve models}
\label{app:clusters}

\begin{figure*}[t]
\centering
\includegraphics[width=\textwidth]{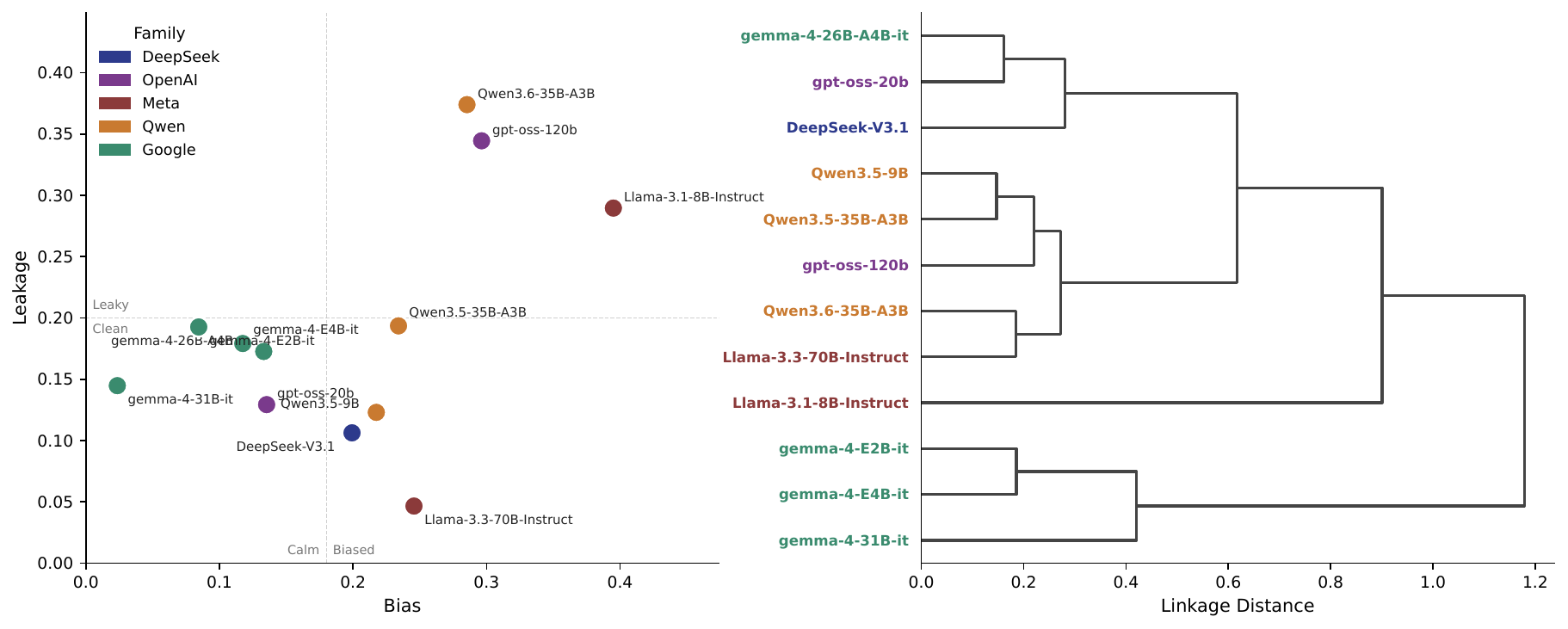}
\caption{Multidimensional clustering of the twelve audited models. \textbf{Left}: scatter of mean role bias $|\hat\beta_p^{\mathrm{PR}}|$ versus mean leakage $L_m$, with each model colored by family. Dashed lines partition the plane into the four quadrants discussed below. \textbf{Right}: Ward hierarchical-clustering dendrogram on the $9$-dimensional PR persona-coefficient vector (Euclidean distance). Leaf labels colored by family. The dendrogram reveals strong within-family clustering for Qwen and Gemma, with Llama-3.1-8B-Instruct sitting alone at the largest height.}
\label{fig:model_clusters}
\end{figure*}

The audit gives us several features per model: aggregate metrics (bias, retention, leakage) and a $9$-dimensional persona-coefficient vector. Figure~\ref{fig:model_clusters} projects the twelve models in two complementary ways. The left panel places each model on the (bias, leakage) plane and reveals four practical quadrants: clean-and-calm (low bias, low leakage), clean-but-biased, leaky-but-calm, and leaky-and-biased. The right panel runs Ward hierarchical clustering on the $9$-dim PR persona vector and shows which models behave most similarly across personas.

\paragraph{Quadrants on the (bias, leakage) plane.} Using split lines at $\overline{|\hat\beta_p^{\mathrm{PR}}|} = 0.18$ and $\overline{L_m} = 0.18$ (the leakage split is the panel median; the bias split sits just below the $0.21$ panel-median magnitude of Table~\ref{tab:crossmodel} to separate the visibly low-bias models):
\begin{itemize}[leftmargin=*,itemsep=1pt,topsep=2pt]
\item \emph{Clean-and-calm} (low bias, low leakage): gemma-4-31B-it, gemma-4-E2B-it, gpt-oss-20b. These models barely respond to the role prompt and route what little they pick up cleanly. Safe for invariance but plausibly under-personalizing.
\item \emph{Clean-but-biased} (high bias, low leakage): Llama-3.3-70B-Instruct, DeepSeek-V3.1, Qwen3.5-9B. Strong role response \emph{and} clean routing. This is the best operating point: the model engages with the role but keeps the user-side signal off the evidence-based field.
\item \emph{Leaky-and-biased} (high bias, high leakage): gpt-oss-120b, Llama-3.1-8B-Instruct, Qwen3.6-35B-A3B. Strong role response but the schema fails to keep the bias out of the evidence-based field. Worst quadrant for deployment under invariance constraints.
\item \emph{Leaky-but-calm} (low bias, high leakage): gemma-4-E4B-it sits near here. Modest bias but proportionally large fraction of the user signal still leaks through.
\end{itemize}

\paragraph{What the dendrogram tells us.} Cutting the tree at $k=4$ recovers four behavior groups (visible as the four color-grouped subtrees on the right panel of Figure~\ref{fig:model_clusters}):

\begin{itemize}[leftmargin=*,itemsep=1pt,topsep=2pt]
\item \emph{Llama-3.1-8B-Instruct} stands alone; it sits at the largest height, indicating its $9$-dim persona response is the farthest from every other model's. This is the same model that has the largest $|\hat\beta_p^{\mathrm{PR}}|$ on ShortSeller ($-0.741$), and it amplifies the persona response more strongly than its peers across the board.
\item \emph{A Qwen-dominated cluster} containing Qwen3.6-35B-A3B, Qwen3.5-35B-A3B, Qwen3.5-9B, plus Llama-3.3-70B-Instruct and gpt-oss-120b. These models share a similar persona-response \emph{shape} despite spanning $9$B to $\sim 120$B active parameters: family identity matters more than scale for this similarity.
\item \emph{A "balanced" cluster} containing DeepSeek-V3.1, gpt-oss-20b, and gemma-4-26B-A4B-it: moderate bias on every persona, no extreme cells.
\item \emph{A small-Gemma cluster} containing gemma-4-31B-it, gemma-4-E4B-it, gemma-4-E2B-it: weak persona response across the board (the cluster characterized by the flat persona row in Figure~\ref{fig:model_persona}).
\end{itemize}

\paragraph{Implication.} On the (bias, leakage) plane the four quadrants give a deployer a one-glance map of which models are safe-for-invariance, safe-for-personalization, both, or neither. The dendrogram says that behavioral similarity does not track family or scale cleanly: a Qwen MoE, an OpenAI MoE, and a Meta dense model can land in the same behavioral cluster. The audit therefore generalizes in a non-obvious way, two models that look unrelated by architecture or training data can behave nearly identically on the persona dimension.

\subsection{Extreme filing cases}
\label{app:extremes}

Table~\ref{tab:extremes} lists the ten filings in the audit sample with the largest cross-persona evidence-based-score spread under PR on DeepSeek-V3.1. These are the filings where the persona-induced bias is most visible at the single-document level. The top spread is $1.45$ units (Cooper Companies $10$-Q); all ten are at least $1.30$. Sector composition: Consumer Discretionary appears three times, Health Care three times, with Energy, InfoTech, Industrials, and Financials each appearing once. Filings in these sectors are most exposed to persona-induced verdict instability.

\begin{table*}[h]
\centering
\small
\setlength{\tabcolsep}{4pt}
% \begin{tabular}{lllrrr}
\begin{tabular*}{\textwidth}{@{\extracolsep{\fill}} lllrrr @{}}
\toprule
Company & Sector & Type & \multicolumn{1}{c}{$\min$} & \multicolumn{1}{c}{$\max$} & spread \\
\midrule
Cooper Companies     & Health      & 10-Q & $-0.60$ & $+0.85$ & $1.45$ \\
Royal Caribbean      & Cons.\ Disc. & 10-Q & $-0.60$ & $+0.80$ & $1.40$ \\
PulteGroup           & Cons.\ Disc. & 10-K & $-0.60$ & $+0.70$ & $1.30$ \\
D.R.\ Horton         & Cons.\ Disc. & 10-Q & $-0.60$ & $+0.70$ & $1.30$ \\
Hess Corp            & Energy      & 10-Q & $-0.60$ & $+0.70$ & $1.30$ \\
KLA Corp             & InfoTech    & 10-K & $-0.60$ & $+0.70$ & $1.30$ \\
Humana               & Health      & 10-Q & $-0.60$ & $+0.70$ & $1.30$ \\
Cooper Companies     & Health      & 10-K & $-0.60$ & $+0.70$ & $1.30$ \\
Expeditors Intl.     & Industrials & 10-K & $-0.60$ & $+0.70$ & $1.30$ \\
SVB Financial Group  & Financials  & 10-Q & $-0.60$ & $+0.70$ & $1.30$ \\
\bottomrule
\end{tabular*}
\caption{Top-10 filings by cross-persona evidence-based-score spread under PR on DeepSeek-V3.1. ``Spread'' = $\max_p \hat s_{i,p}^{\text{ev}} - \min_p \hat s_{i,p}^{\text{ev}}$ across the nine non-neutral personas on filing $i$.}
\label{tab:extremes}
\end{table*}

\subsection{Per-call dual-output scatter}

\begin{figure*}[t]
\centering
\includegraphics[width=\textwidth]{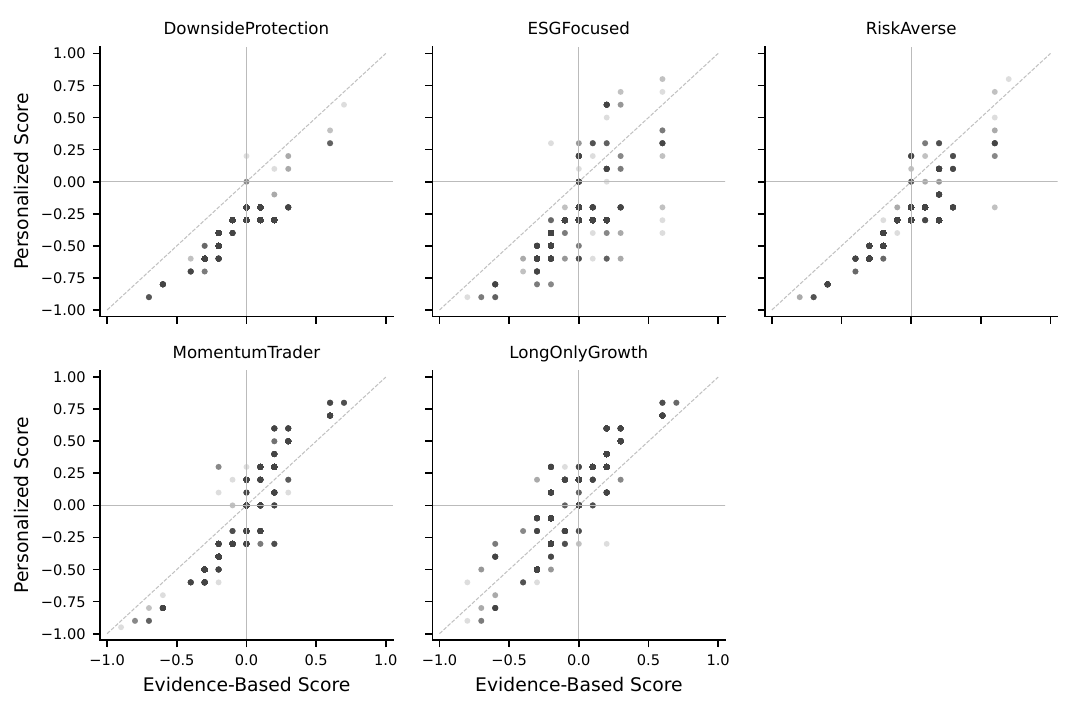}
\caption{Per-call scatter of evidence-based score (x-axis) versus personalized score (y-axis) under MEM, $n=3{,}575$ per profile. Diagonal: perfect leakage. Y-axis: perfect routing. Origin: over-suppression.}
\label{fig:dual_scatter}
\end{figure*}

Figure~\ref{fig:dual_scatter} shows the per-call placement of the evidence-based and personalized scores on the MEM arm, faceted by user profile. Clean routing puts mass on the y-axis (large personalized, small evidence-based); leakage puts mass on the diagonal. Downside protection and Risk-averse clouds extend down-and-leftward (bearish personalized score $+$ small but non-zero negative evidence-based score), and the leakage observed at the coefficient level in Table~\ref{tab:crossmodel} is visible at the call level. ESG-focused mass clusters near the y-axis (cleanest routing). Momentum trader mass clusters near the origin (over-suppression).

\subsection{Forward returns and forward volatility by score quintile}

A natural follow-up question is whether the role-conditioned evidence-based score carries any ex-post signal about the firm's future performance, i.e., whether the bias is just rhetoric, or whether the within-persona cross-section of scores has predictive content. For each (model, persona) we sort the $\sim 3{,}393$ filings (the subset with at least $252$ post-filing trading days) into score quintiles (random tie-breaking on the coarse score grid) and compute the mean $252$-trading-day cumulative abnormal return on the top minus bottom quintile (Q5$-$Q1). The signal is reported two ways: per persona across the three biggest models in the panel and a 12-model panel mean (Table~\ref{tab:quintile_by_persona}); and per model averaged over the nine personas (Table~\ref{tab:quintile_by_model}).

\begin{figure}[t]
\centering
\includegraphics[width=\columnwidth]{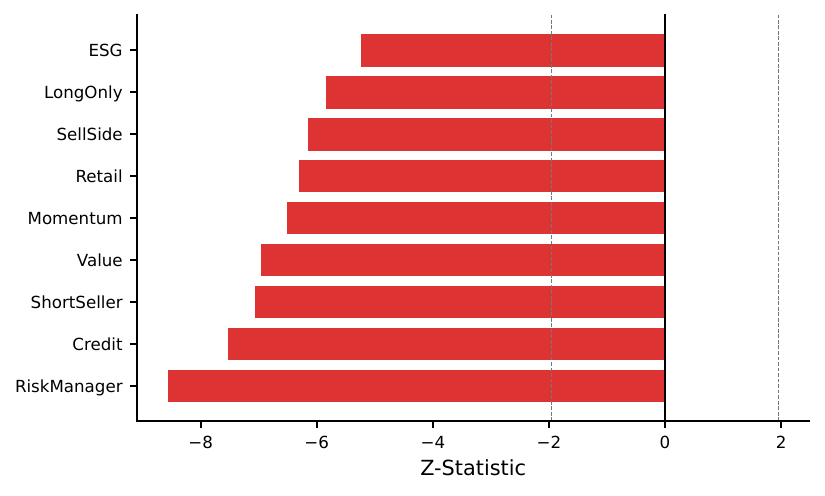}
\caption{Persona-level $z$-statistics on the evidence-based-score coefficient in an OLS regression of $\log$ forward $21$-day realized volatility on the score, with controls (prior $21$-day abnormal return, $90$-day market beta, sector fixed effects, HC1 standard errors). Dashed lines: $z = \pm 1.96$.}
\label{fig:vol_zstats}
\end{figure}

\paragraph{Cross-model patterns.}
Three observations from Table~\ref{tab:quintile_by_persona}. \emph{(i) The within-persona signal is robust across models}: every persona produces a non-negative Q5$-$Q1 on the 12-model panel mean (Risk manager only marginally, at $+0.08\%$), and the average across personas is $+3.12\%$ (DeepSeek), $+3.17\%$ (gpt-oss-120b), and $+2.05\%$ (Llama-3.3-70B-Instruct), all economically meaningful at the $252$-day horizon. \emph{(ii) Risk manager is universally inverted}: the Q5$-$Q1 spread is negative on the three biggest models and very near zero on the panel mean. The DeepSeek finding ($-1.40\%$) is not a model-specific anomaly but a systematic feature of the persona itself, ``less bad'' filings under a risk-management read underperform ``more bad'' filings, while the same score still predicts forward volatility correctly (Figure~\ref{fig:vol_zstats}). \emph{(iii) Cross-persona ordering is model-specific}: Credit is DeepSeek's strongest persona ($+5.30\%$) but Llama-3.3-70B-Instruct's is Value ($+5.94\%$); the per-persona rankings do not commute across models.

\begin{table}[t]
\centering
% \small
\setlength{\tabcolsep}{6pt}
\begin{tabular*}{\columnwidth}{@{\extracolsep{\fill}} lr @{}}
\toprule
Model & $\overline{\text{Q5}{-}\text{Q1}}$ \\
\midrule
Qwen3.5-35B-A3B      & $+6.35\%$ \\
Qwen3.6-35B-A3B      & $+5.76\%$ \\
gemma-4-31B-it              & $+4.66\%$ \\
Qwen3.5-9B               & $+4.05\%$ \\
gemma-4-26B-A4B-it          & $+3.32\%$ \\
gpt-oss-120b             & $+3.17\%$ \\
DeepSeek-V3.1            & $+3.12\%$ \\
Llama-3.3-70B-Instruct            & $+2.05\%$ \\
gemma-4-E4B-it          & $+1.64\%$ \\
gemma-4-E2B-it              & $+0.69\%$ \\
gpt-oss-20b              & $+0.39\%$ \\
Llama-3.1-8B-Instruct             & $+0.32\%$ \\
\midrule
\textbf{Panel mean}      & $\mathbf{+2.96\%}$ \\
\bottomrule
\end{tabular*}
\caption{Per-model Q5$-$Q1 cumulative abnormal return at $252$ days, averaged over the nine personas, sorted descending. Bigger models do not produce stronger signal: Qwen3.5-35B-A3B leads at $+6.35\%$ and the $\sim 685$B DeepSeek-V3.1 is mid-pack at $+3.12\%$, while the three smallest dense or MoE models (Llama-3.1-8B-Instruct, gpt-oss-20b, gemma-4-E2B-it) produce essentially zero signal. The cross-section of scores carries predictive content even from models whose absolute bias is low (Qwen3.5-35B) and from models whose absolute bias is high (Llama-3.1-8B-Instruct at $0.40$ panel-mean bias), but the magnitude of the signal does not track the magnitude of the bias.}
\label{tab:quintile_by_model}
\end{table}

\begin{figure}[t]
\centering
\includegraphics[width=\columnwidth]{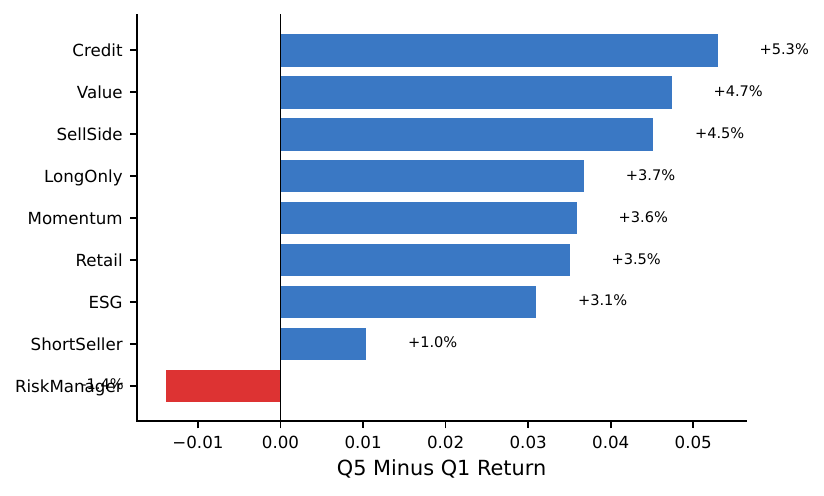}
\caption{Q5$-$Q1 cumulative abnormal return at $252$ days, by persona, on DeepSeek-V3.1 (the headline model). Numerical values in column 1 of Table~\ref{tab:quintile_by_persona}.}
\label{fig:returns_quintile}
\end{figure}

We also estimate, persona-by-persona, an OLS regression of log forward $21$-day realized volatility on the evidence-based score, with controls for prior $21$-day abnormal return, $90$-day market beta, and sector fixed effects (HC1 standard errors). The $z$-statistic on the score coefficient ranges from $-5.3$ to $-8.6$ across the nine personas (Figure~\ref{fig:vol_zstats}): higher (more bullish) score predicts lower realized volatility on every persona. Risk manager carries the strongest signal here ($z = -8.6$, $\hat\beta = -0.45$) despite producing an inverted return signal in Table~\ref{tab:quintile_by_persona}, consistent with the model's risk-manager judgment being well-calibrated on volatility but mis-calibrated on direction.

The implication for the paper's broader claim is twofold. First, role-conditioned bias is not purely cosmetic: within a persona, the cross-section of scores has substantive predictive content for both forward returns and forward volatility. Second, the bias \emph{direction} matters: bearish role conditioning shifts the level of the score (the panel-mean coefficient we report in Section~\ref{sec:evidence_selection}), not the within-persona ranking. The cross-section retains its signal even when the level is biased. This is consistent with the audit-design separation in the main text: what changes across personas is largely the framing of identical evidence, not the model's relative ranking of evidence.

\subsection{Worked example: a sign flip on the same filing}
\label{app:case_study}

To make the channel decomposition concrete at the call level, we walk through one filing where the cross-persona disagreement is large enough to span both signs of the score range. We pick the filing with the largest LongOnly--ShortSeller gap under PR out of the $3{,}575$ in the sample.

\paragraph{The filing.} The Coca-Cola Company, Form 10-Q, filed $2023$-$10$-$24$ (CIK $21344$, accession \texttt{0000021344-23-000060}). The disclosure contains both bullish material (\$$15.4$B in cash and equivalents; strong long-term borrowing capacity) and bearish material (an active IRS tax litigation estimated at \$$5.6$B for the $2007$--$2009$ tax years plus \$$1.2$B incremental for the first nine months of $2023$, and exposure to OECD Pillar Two global minimum tax). Reasonable analysts would disagree on the net read; the audit asks whether the model's read is itself unstable as the user context changes.

\paragraph{Retrieved evidence (NR and MEM arms, byte-identical $k=8$).} The neutral retrieval query returns the same eight chunks for every persona under NR and every user-memory profile under MEM:

\begin{Verbatim}
risk_factors::0000, mdna::0000, quant_qual::0000,
mdna::0045, mdna::0046, mdna::0047, mdna::0049, mdna::0054
\end{Verbatim}

The three most decision-relevant chunks (verbatim excerpts; ellipses for whitespace):

\begin{Verbatim}
[chunk_id=...::mdna::0046]
LIQUIDITY, CAPITAL RESOURCES AND FINANCIAL POSITION
We believe our ability to generate cash flows from operating
activities is one of the fundamental strengths of our business.
... We have a history of borrowing funds both domestically and
internationally at reasonable interest rates ... The Company's
cash, cash equivalents, short-term investments and marketable
securities totaled $15.4 billion as of September 29, 2023.

[chunk_id=...::mdna::0049]
The Company estimates the impact of the continued application of
the Tax Court Methodology for the nine months ended September 29,
2023 would increase the potential aggregate incremental tax and
interest liability by approximately $1.2 billion. Once the Tax
Court renders a decision, the Company will have 90 days to file a
notice of appeal and pay the portion of the potential aggregate
incremental tax and interest liability related to the 2007 through
2009 tax years, which we currently estimate to be approximately
$5.6 billion (including interest accrued through September 29,
2023) ... While we believe it is more likely than not that we will
prevail in the tax litigation ...

[chunk_id=...::mdna::0045]
... Model rules adopted pursuant to this [OECD] project would
establish a global per-country minimum tax of 15%, and the
European Union has approved a directive requiring member states to
incorporate similar provisions into their respective domestic laws.
... Our income tax expense could be unfavorably impacted as the
legislation becomes effective in countries in which we do business.
\end{Verbatim}

\paragraph{What each arm returns.} Table~\ref{tab:case_study} shows the per-call scores and recommendations under PR, NR, and MEM, for the two extreme personas and their matched user-memory profiles plus the neutral baseline.

\begin{table}[t]
\centering
\small
\setlength{\tabcolsep}{3pt}
\begin{tabular}{llrrl}
% \begin{tabular*}{\columnwidth}{@{\extracolsep{\fill}} {llrrl} @{}}
\toprule
Arm & Persona / profile & $\hat s^{\text{ev}}$ & $\hat s^{\text{pers}}$ & Reco \\
\midrule
\multirow{3}{*}{PR}  & LongOnly             & $+0.60$ & $+0.70$ & BUY  \\
                     & Neutral persona      & $+0.20$ & $+0.10$ & HOLD \\
                     & ShortSeller          & $-0.60$ & $-0.80$ & SELL \\
\midrule
\multirow{3}{*}{NR}  & LongOnly             & $+0.20$ & $+0.30$ & BUY  \\
                     & Neutral persona      & $+0.20$ & $+0.10$ & HOLD \\
                     & ShortSeller          & $-0.30$ & $-0.70$ & SELL \\
\midrule
\multirow{3}{*}{MEM} & Long-only growth     & $+0.20$ & $-0.30$ & HOLD \\
                     & Neutral user profile & $+0.20$ & $+0.10$ & HOLD \\
                     & Downside protection  & $+0.10$ & $-0.30$ & HOLD \\
\bottomrule
\end{tabular}
\caption{Coca-Cola 10-Q $2023$-$10$-$24$ scored under all three conditions. $\hat s^{\text{ev}}$ is the evidence-based score field; $\hat s^{\text{pers}}$ is the personalized score field. NR and MEM share byte-identical retrieved chunks.}
\label{tab:case_study}
\end{table}

\paragraph{Reading the table.} Three observations.

\emph{(1) The model retrieves different chunks under different personas in PR.} The LongOnly PR call retrieves \texttt{mdna::0046} (the \$$15.4$B liquidity chunk) as its second-highest hit; the ShortSeller PR call retrieves it ninth (it falls outside the top $k=8$). This is the evidence-selection channel: the bearish persona is reading a partly different filing.

\emph{(2) The sign flip persists under byte-identical evidence.} Under NR both personas read the same eight chunks, including \emph{both} the bullish \texttt{mdna::0046} liquidity chunk \emph{and} the bearish \texttt{mdna::0049} tax-litigation chunk. LongOnly produces $\hat s^{\text{ev}} = +0.20$ with a BUY recommendation and a rationale built around the liquidity number; ShortSeller produces $\hat s^{\text{ev}} = -0.30$ with a SELL recommendation and a rationale built around the \$$5.6$B contingent liability. The NR spread $|\hat s^{\text{ev}}_{\text{long}} - \hat s^{\text{ev}}_{\text{short}}| = 0.50$ is $42\%$ of the PR spread of $1.20$, so the role-interpretation channel accounts for the larger part of the disagreement on this filing.

\emph{(3) The dual-output schema (MEM) routes the disagreement out of the evidence-based field.} Reframing the same investor mindset as a user-profile served by a neutral assistant collapses the evidence-based scores into a $[+0.10, +0.20]$ band across all three meaningful profiles (Long-only growth, neutral user profile, Downside protection). The personalized field still carries the bearish read for Downside protection ($-0.30$) and \emph{still carries a bearish read} for Long-only growth ($-0.30$ as well); the model interprets a long-only growth user as being more cautious about a \$$5.6$B tax overhang than the role-prompted LongOnly persona was. The personalized \emph{recommendation} is HOLD across all MEM cells. The evidence-based field, which we instructed the model to keep user-invariant, is approximately user-invariant in MEM but not in NR.

\paragraph{What the rationales look like.} The verbatim model outputs are short enough to quote. Under NR / ShortSeller:

\begin{small}
\textit{``As an activist short-seller, the disclosed tax litigation and OECD tax risks are major red flags. The \$$5.6$ billion potential liability is material relative to net income, and management's confidence may be overoptimistic. The $120$-day supplier payment terms and reliance on debt financing also suggest potential cash flow strain if tax payments become due.''}
\end{small}

Under NR / LongOnly:

\begin{small}
\textit{``As a long-only investor seeking compounders, Coca-Cola's strong liquidity, cash flow generation, and commitment to dividends and share repurchases align with your focus on durable strengths. The tax litigation is a known risk but manageable given the company's financial position, and the filing does not reveal any fundamental deterioration in the business.''}
\end{small}

Under MEM / Downside protection:

\begin{small}
\textit{``Given the user's tail-risk focus, the \$$5.6$ billion tax liability and OECD tax reform uncertainties are significant adverse disclosures that increase downside risk. While liquidity appears strong, these legal and regulatory risks warrant a cautious stance, leading to a HOLD rather than BUY.''}
\end{small}

The MEM rationale acknowledges both sides of the disclosure (``liquidity appears strong'' and ``\$$5.6$ billion tax liability''), explicitly attributes the cautious framing to the user's profile (``given the user's tail-risk focus''), and resolves to HOLD rather than SELL. The NR / ShortSeller rationale, in contrast, treats management's confidence as ``overoptimistic'' and produces a SELL. Same evidence, same model, different framing: same factual content marshalled toward opposite verdicts.

\subsection{Per-condition completion}

Invalid or non-conforming outputs were excluded from the per-filing regression. The panel regressions use the complete-intersection set across all three conditions.

\subsection{Models agree on which roles move judgments}
\label{sec:model_agreement}

\begin{figure*}[!t]
\centering
\includegraphics[width=\textwidth]{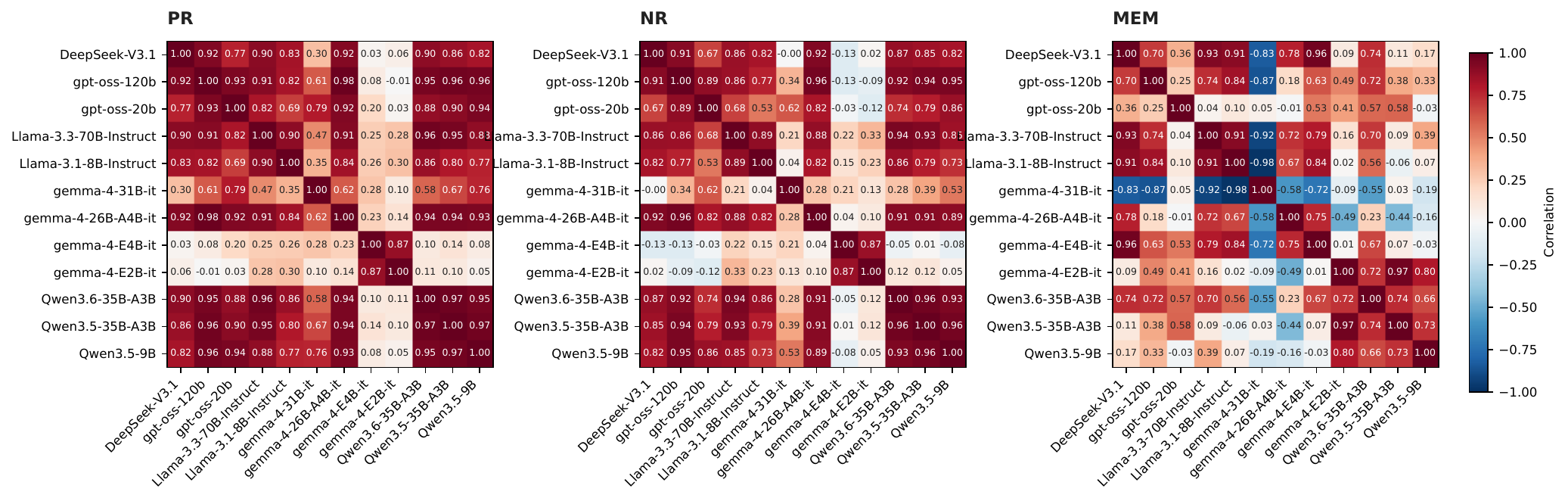}
\caption{Model agreement on per-persona patterns. From left to right: PR, NR, MEM. Each cell shows the Pearson correlation between two models' coefficient vectors on the \textbf{evidence-based} score: $9$-dimensional persona vectors under PR and NR, and $5$-dimensional profile vectors under MEM. Positive correlation indicates agreement on which personas (or profiles) are bearish or bullish.}
\label{fig:model_corr}
\end{figure*}

The effects are structured rather than random. We represent each model by its vector of persona coefficients and compare these vectors across models. Under PR and NR, many models agree on the relative ordering of roles: ShortSeller and Risk manager tend to push the evidence-based score downward, while LongOnly tends to push it upward. This agreement persists even when magnitudes differ, suggesting that role prompts induce a shared bearish/bullish structure rather than idiosyncratic noise. Under MEM, correlations are less uniform, consistent with the smaller evidence-side coefficients and with model-specific differences in how well user-profile information is routed away from the neutral score (Figure~\ref{fig:model_corr}).

\subsection{Cross-domain sanity check: GovReport, MS\textsuperscript{2}, CAP}
\label{app:cross_domain}

\paragraph{Setup.}
We re-run a reduced version of the audit on three non-finance corpora chosen to test whether the user-context bias channel and the schema-based mitigation generalize outside SEC filings. Each corpus is a long-form synthesis task with an anchored neutral baseline and stakeholder-emphasis personas (different priorities on shared evidence, not disputes about the evidence itself). DeepSeek-V3.1 only; deterministic decoding ($T{=}0$); the evidence is supplied directly so the PR arm is absent and we report only the NR$\to$MEM (role $\to$ user-profile) comparison. The dual-output schema mirrors the finance schema---two $[-1,+1]$ score fields plus a categorical recommendation---except that the \texttt{personalized\_recommendation} field uses task-appropriate \texttt{SUPPORTS}/\texttt{INCONCLUSIVE}/\texttt{REFUTES} labels in place of \texttt{BUY}/\texttt{HOLD}/\texttt{SELL}.

\begin{itemize}[leftmargin=*,itemsep=1pt,topsep=2pt]
\item \textbf{GovReport} \citep{huang2021govreport}: CRS/GAO policy reports vs.\ their expert summaries (median $6{,}620$ words evidence, $n=500$). Personas: Fiscal hawk, Public-interest analyst, Implementation-risk analyst.
\item \textbf{MS\textsuperscript{2}} \citep{deyoung2021ms2}: bodies of medical-trial abstracts vs.\ a Cochrane-style synthesis (median $5{,}765$ words evidence, $n=200$). Personas: Clinician, Payer, Patient advocate.
\item \textbf{CAP} \citep{harvardlilcoldcases}: court opinions vs.\ the constructed proposition that ``the opinion's reasoning supports the holding it reaches'' (median $2{,}673$ words evidence, $n=200$). Personas: Judicial clerk, Trial practitioner, Legal scholar.
\end{itemize}

In all three settings the gold label is SUPPORTS by construction (the summary, synthesis, or holding is the document's own canonical conclusion), so any movement off SUPPORTS under role/profile framing is bias, not genuine disagreement with the evidence.

\paragraph{Results.}
Per-cell shifts and shrinkage are reported in Table~\ref{tab:cross_domain_summary}.

\begin{table}[t]
\centering
\small
\setlength{\tabcolsep}{5pt}
\resizebox{\columnwidth}{!}{%
\begin{tabular}{lrrr}
\toprule
& $|\beta_{NR}|$ & $|\gamma_{MEM}|$ & shrinkage \\
\midrule
\multicolumn{4}{l}{\textbf{Finance} ($n=3{,}575$; main audit, DeepSeek-V3.1)} \\
ShortSeller / Downside protection   & $0.195$ & $0.045$ & $+77\%$ \\
Risk manager / Risk-averse         & $0.266$ & $0.058$ & $+78\%$ \\
LongOnly / Long-only growth        & $0.136$ & $0.013$ & $+91\%$ \\
ESG / ESG-focused                  & $0.044$ & $0.010$ & $+77\%$ \\
Momentum / Momentum trader         & $0.025$ & $0.004$ & $+85\%$ \\
\textit{Finance median}           & $\mathit{0.14}$  & $\mathit{0.013}$ & $\mathbf{+80\%}$ \\
\midrule
\multicolumn{4}{l}{\textbf{GovReport} ($n=500$)} \\
Fiscal hawk                       & $0.024$ & $0.016$ & $+32\%$ \\
Public-interest analyst           & $0.017$ & $0.013$ & $+24\%$ \\
Implementation-risk analyst       & $0.024$ & $0.014$ & $+40\%$ \\
\textit{GovReport median}         & $\mathit{0.024}$ & $\mathit{0.014}$ & $\mathbf{+32\%}$ \\
\midrule
\multicolumn{4}{l}{\textbf{MS\textsuperscript{2}} ($n=200$)} \\
Clinician                         & $0.100$ & $0.097$ & $+3\%$  \\
Payer                             & $0.140$ & $0.109$ & $+22\%$ \\
Patient advocate                  & $0.114$ & $0.101$ & $+11\%$ \\
\textit{MS\textsuperscript{2} median}
                                  & $\mathit{0.114}$ & $\mathit{0.101}$ & $\mathbf{+11\%}$ \\
\midrule
\multicolumn{4}{l}{\textbf{CAP} ($n=200$)} \\
Judicial clerk                    & $0.012$ & $0.013$ & $-8\%$  \\
Trial practitioner                & $0.012$ & $0.011$ & $+6\%$  \\
Legal scholar                     & $0.034$ & $0.019$ & $+42\%$ \\
\textit{CAP median}               & $\mathit{0.012}$ & $\mathit{0.013}$ & $\mathbf{+6\%}$  \\
\bottomrule
\end{tabular}%
}
\caption{Per-cell NR$\to$MEM shrinkage on the EvidenceScore on DeepSeek-V3.1. Finance rows are the five matched pairs from the main audit (Table~\ref{tab:full_bias}). The three cross-domain blocks each report three stakeholder-emphasis personas on a long-form synthesis task with an anchored neutral baseline. All shrinkages are paired differences against an example-level neutral call.}
\label{tab:cross_domain_summary}
\end{table}

\paragraph{Shrinkage tracks the room the task gives the bias channel.}
The cross-domain results are most usefully read together rather than dataset-by-dataset. Finance has, by some distance, the largest underlying bias channel: median $|\hat\beta_p^{\mathrm{NR}}|{=}0.14$, with individual matched pairs running as high as $0.27$. This is not surprising. Reading an SEC 10-K for an investment outlook is a synthesis task whose ``correct'' answer is genuinely subjective, a short-seller and a long-only investor reading the same disclosure can reach different conclusions without either being factually wrong, so the persona prompt has a lot of interpretive room to work in. The three non-finance corpora are all more anchored: a CRS report summarizes itself, a Cochrane-style synthesis is meant to be the canonical reading of its input studies, and a court opinion is internally written to support the holding the court reaches. The bias channel on these documents is therefore much smaller: GovReport median $|\hat\beta_p^{\mathrm{NR}}|{=}0.024$, MS\textsuperscript{2} $0.114$, CAP $0.012$, roughly $6\times$, $1.2\times$, and $12\times$ smaller than finance, respectively. Cross-domain shrinkage in absolute terms is correspondingly smaller: GovReport median $+32\%$, MS\textsuperscript{2} $+11\%$, CAP $+6\%$. Crucially, the direction of the schema's response is consistent: $8$ of the $9$ cross-domain cells show non-negative shrinkage (the schema reduces or at worst leaves unchanged the persona signal in the evidence-based score), and the one negative cell (CAP / Judicial clerk at $-8\%$) sits in the regime where $|\hat\beta_p^{\mathrm{NR}}|$ is so small that the shrinkage ratio is dominated by paired noise. The mechanism the main audit identifies on finance, user-profile framing routes user signal off the supposedly persona-invariant field, thus appears to replicate on three independent non-finance long-form synthesis tasks, at magnitudes set by how much interpretive room the document itself leaves available to be biased in the first place.

\end{document}